\documentclass{article}
\usepackage{amssymb}

\usepackage[eandd, final]{neurips_2026}
\usepackage{amssymb}

\usepackage[T1]{fontenc}
\usepackage[utf8]{inputenc}
\usepackage{microtype}
\usepackage{graphicx}
\usepackage{booktabs}
\usepackage{array}
\usepackage{amsmath}
\usepackage{enumitem}
\usepackage{xcolor}
\usepackage{hyperref}
\usepackage{float}
\usepackage{subcaption}
\usepackage{multirow}
\usepackage{nicefrac}
\usepackage{wrapfig}

\newcommand{\pless}{P-less}

\title{Long-Context Generation Is a Sampling Problem\\(When You Turn Up the Heat)}

\author{%
  Allen G.\ Roush \\
  Thoughtworks \\
  \texttt{allen.roush@thoughtworks.com} \\
  \And
  Nguyen Nhat Minh \\
  Apart Research \\
  \texttt{minh1228@gmail.com} \\
  \And
  Ravid Shwartz-Ziv \\
  New York University \& Meta \\
  \texttt{ravid.shwartz.ziv@nyu.edu} \\
  \AND
  Judah Goldfeder \\
  Columbia University \\
  \texttt{jag2396@columbia.edu} \\
  \And
  Sanjay Basu \\
  Oracle \\
  \texttt{sanjay.basu@oracle.com} \\
}

\begin{document}
\maketitle

\begin{abstract}
Open-weight language models are widely observed to get worse the longer they generate. Often, attempts to fix this involve more training. At high temperature, The sampler, i.e. the rule that picks the next token out of the model's probabilities, decides whether long and agentic generation survives and/or thrives. We test 29 open-weight models from 1B to 122B parameters on ten established benchmarks plus needle retrieval out to one million tokens, covering SWE-bench Verified coding agents, long chain-of-thought reasoning, long-form writing, long structured output and long-context reading. Samplers that keep their pool of candidate tokens small and steady as temperature rises lose little or nothing. Samplers that let in low-probability tokens fall apart. At temperature 2, coding agents running top-$p$, top-$k$ or Top-H resolve 0 of 100 GitHub issues, while the same agents running top-$n\sigma$ resolve 46 to 62 and win every issue the two samplers disagree on, on three Qwen3.5 models. Over all benchmarks top-$n\sigma$ keeps a median 100\% of its temperature-1 score where top-$p$ keeps 26\%. Counting how many tokens each sampler lets through explains why. In one Llama-3.1-8B story, going from temperature 1 to 2 grows top-$p$'s pool from 4 tokens to 51,845 while top-$n\sigma$ stays at 3. Across 133 combinations of sampler setting and temperature, the typical pool size predicts judged quality (Spearman correlation of $-0.77$), and when we pick the pool at temperature 1 and only apply temperature afterwards, Top-H, min-$p$ and P-less stop collapsing. At the default temperature of 1 most samplers perform about the same, We release all code and data \href{https://github.com/Hellisotherpeople/Long-Context-Generation-Is-a-Sampling-Problem-8-}{here}.
\end{abstract}

\section{Introduction}

Open-weight language models now generate very long sequences~\citep{dubey2024llama3,yang2024qwen2}. In practice, on smaller models, the second half of a long sample often turns into repeated phrases and loops~\citep{holtzman2020curious,welleck2020neural,fu2021theoretical}, and most proposed fixes involve retraining or long-output fine-tuning~\citep{bai2025longwriter}. We argue for a cheaper fix that inference engines already expose, which is the sampler.

\paragraph{Samplers and temperature.}
At each step the model gives a probability for every token in its vocabulary, and the sampler stacks outputs a single choice. Given a temperature of 1 or below, it piles probability onto the likeliest tokens, and above 1 it spreads probability out toward unlikely ones. Truncation rules cut the unlikely tokens and samples from the remaining distribution, which we call the \emph{candidate set}. Top-$k$ keeps a fixed number of tokens~\citep{fan2018hierarchical}. Top-$p$ keeps however many tokens it takes to cover a fixed share of the probability~\citep{holtzman2020curious}. Newer rules like min-$p$~\citep{nguyen2024minp}, top-$n\sigma$~\citep{tang2024topnsigma}, Top-H~\citep{baghaei2025toph} and P-less~\citep{tan2025pless} look at the shape of the distribution at each step before deciding where to cut.

\begin{figure}[t]
\centering
\includegraphics[width=\textwidth]{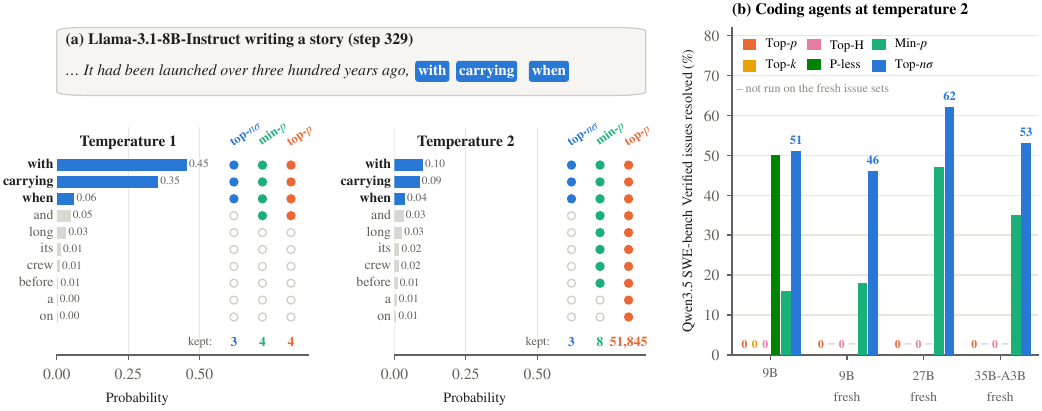}
\caption{\textbf{(a)} Llama-3.1-8B-Instruct partway through a science-fiction story (the step is picked by a fixed rule, Appendix~\ref{app:numbers}). The box shows the text so far, with the three continuations top-$n\sigma$ keeps highlighted. The bars are the model's ten likeliest next tokens, each dot column marks the tokens one rule keeps, and the number under a column is that rule's total. Going from temperature 1 to 2 flattens the bars. Top-$n\sigma$ still keeps the same three words and min-$p$ only grows from 4 to 8, but top-$p$ 0.9 now needs 51,845 tokens, most of the 128K vocabulary, to reach 90\% of the probability. \textbf{(b)} At temperature 2, Qwen3.5 agents using top-$p$, top-$k$ or Top-H resolve no SWE-bench Verified issues, while top-$n\sigma$ agents resolve 46--62\% (100 issues per group, dashes mark samplers that were not run on the fresh issue sets).}
\label{fig:teaser}
\end{figure}

\section{Background and Related Work}
\label{sec:related}

\paragraph{Truncation samplers.}
Other rules besides the ones above include locally typical sampling~\citep{meister2023locally}, $\eta$-sampling~\citep{hewitt2022truncation}, Mirostat~\citep{basu2021mirostat} and XTC~\citep{weidmann2024xtc,weidmann2026xtc}. Top-$n\sigma$, Top-H and P-less all set their cutoff from statistics of the whole distribution, which is why we have lumped them together as ``full-distribution-aware''.

\paragraph{Degeneration.}
It has been known for years that neural text generators fall into repetition over long outputs~\citep{holtzman2020curious,welleck2020neural,fu2021theoretical}. Repetition penalties~\citep{keskar2019ctrl}, diversity objectives~\citep{li2016diversity}, the DRY penalty~\citep{weidmann2024dry,weidmann2026dry} and contrastive search~\citep{su2022contrastive,su2023contrastive} were all built to deal with it. We focus on the base truncation rule and look at penalties separately in Appendix~\ref{app:extra}.

\paragraph{Evaluating samplers and long context.}
\citet{larosa2026fragility} ran a controlled thirteen-model study and found that truncation samplers mostly change short-answer accuracy when high temperature has already degraded the model. We find that this regime also governs long-context and agentic generation. Long-context ability is usually measured with retrieval and comprehension benchmarks~\citep{bai2024longbench,hsieh2024ruler,shaham2023zeroscrolls}, and long-output ability with benchmarks like LongBench-Write~\citep{bai2025longwriter}. LLM judges~\citep{zheng2023judging,li2024generative} make it practical to grade long outputs one window at a time.

\section{The Impact of Temperature On Truncation Samplers at Long Context}
\label{sec:method}

Table~\ref{tab:glance} describes the truncations rules we compare in plain terms. Formal definitions are in Appendix~\ref{app:defs}. HuggingFace, vLLM and SGLang all apply temperature before truncation. Notably, Llama.cpp choose to default to the opposite order. That ordering matters. With temperature first, going from temperature 1 to 2 grows the median top-$p$ set from 2 tokens to about 90,000, Top-H from 1 to 54, P-less from 1 to 21 and min-$p$ from 1 to 14, because each rule now draws its cutoff on probabilities that temperature has already flattened. Truncate on the untempered probabilities and apply temperature afterwards, and suddenly every rule keeps its temperature-1 set at any temperature. Temperature then only moves probability around among tokens the rule already accepted. Section~\ref{sec:ablation} tests whether this reordering alone fixes long context generation problems.

\begin{table}[t]
\centering
\caption{The samplers we compare, what each one keeps, and how many tokens it keeps per decoding step (median over the first 2K tokens of the ten-model generations of Section~\ref{sec:r1}, full numbers in Table~\ref{tab:pool}). Temperature is applied before truncation, as HuggingFace, vLLM and SGLang do by default. Compare the two columns of each row. Top-$p$, Top-H, P-less and min-$p$ all grow between temperature 1 and 2, and top-$n\sigma$ is the only rule that stays at its temperature-1 size.}
\label{tab:glance}
\small
\setlength{\tabcolsep}{4pt}
\begin{tabular}{l>{\raggedright\arraybackslash}p{6.4cm}cc}
\toprule
 & & \multicolumn{2}{c}{Median tokens kept per step} \\
\cmidrule(lr){3-4}
Sampler & Keeps the tokens that\ldots & temp.\ 1 & temp.\ 2 \\
\midrule
Pure sampling & (all tokens) & all & all \\
Top-$k$ 50 & are among the 50 most likely & 51 & 51 \\
Top-$p$ 0.9 & together cover 90\% of the probability & 2 & $\approx$90{,}000 \\
Top-H 0.4 & keep the set's entropy under 40\% of the full entropy & 1 & 54 \\
Min-$p$ 0.1 & have at least 10\% of the top token's probability & 1 & 14 \\
P-less & are at least as likely as two random draws are to match & 1 & 21 \\
Top-$n\sigma$ 1 & have logits within one standard deviation of the top & 1 & \textbf{1} \\
\bottomrule
\end{tabular}
\end{table}

Top-$n\sigma$ keeps the tokens whose logit sits within $n$ standard deviations of the highest logit. Temperature divides all the logits by the same number, which shrinks the distance to the top and the standard deviation by the same amount, so the tokens top-$n\sigma$ keeps never change with temperature~\citep{tang2024topnsigma}. Temperature only changes how probability gets shared out inside that set. Top-$k$ is fixed by design too, although it hands the model all $k$ tokens even when the model is close to certain. Every other rule sets its cutoff on probabilities that temperature has already flattened, so its set grows as temperature goes up. Min-$p$ and P-less grow slowly because their cutoffs follow the top of the distribution. Top-$p$ and Top-H grow fast because theirs follow cumulative probability and entropy. Top-H reads a global statistic just like top-$n\sigma$ does, yet at high temperature it behaves like top-$p$ for this reason. The low-probability \emph{tail} these fast-growing rules let in is what high temperature turns into noise.

\section{Experimental Setup}
\label{sec:setup}

\paragraph{Models and implementation.}
We use 29 instruction-tuned open-weight models from 1B to 122B parameters from the Llama, Qwen, Mistral, Gemma, OLMo, Nemotron, GLM and gpt-oss families~\citep{dubey2024llama3,yang2024qwen2,jiang2023mistral,openai2025gptoss,xu2025ultralong}, with a core set of ten from Llama-3.2-1B to gpt-oss-120b (Appendix~\ref{app:models} lists all of them). We run the released weights (bf16, or native MXFP4 for gpt-oss-120b) on NVIDIA H200 GPUs. All samplers live in a single vLLM logits processor that replaces native sampling. We checked each one against its authors' reference code, and the processor logs the candidate-set size at every step.

\paragraph{Benchmarks and protocols.}
For agentic coding we use SWE-bench Verified~\citep{jimenez2024swebench} with the mini-swe-agent harness~\citep{minisweagent,yang2024sweagent} and the official test harness. For long chain-of-thought reasoning we use LiveCodeBench v6~\citep{jain2025livecodebench}, AIME and HMMT 2025--26 from MathArena~\citep{balunovic2025matharena} and the level-5 problems of MATH-500~\citep{lightman2024verify}, all in thinking mode with a 32K-token budget. Long-form writing is covered by LongBench-Write~\citep{bai2025longwriter}, WritingBench~\citep{wu2025writingbench} and the 75 English Arena-Write prompts~\citep{wu2026longwriterzero} that ask for at least 2,000 words, all with natural stopping. For long structured output we use LongProc~\citep{ye2025longproc} (six tasks, outputs up to 8K tokens, official rule-based scoring). For long-context reading we use LongBench v2~\citep{bai2025longbenchv2} (503 questions, contexts up to 120K tokens, official chain-of-thought protocol) and multi-key needle retrieval in PG-19 text~\citep{rae2019compressive} from 8K to one million tokens. We also run a stress test that forces eight fiction and eight non-fiction prompts (Appendix~\ref{app:prompts}) out to 16K tokens with end-of-sequence turned off.

\paragraph{Outcomes.}
The LLM judge (gpt-oss-120b) gives a 1 to 10 score to 1.5K-token windows taken at fixed depths. We say a generation has a \emph{terminal loop} if, from some 1K-token window on, at least 80\% of the remaining windows repeat 32-grams at least half the time. A generation is \emph{healthy} if it has no terminal loop and every judged window scores at least 6 for coherence. Coding agents don't need a judge, since issues only count as resolved if the agent's patch passes the repository's tests. 

\section{Results}
\label{sec:results}

\begin{figure}[t]
\centering
\includegraphics[width=\textwidth]{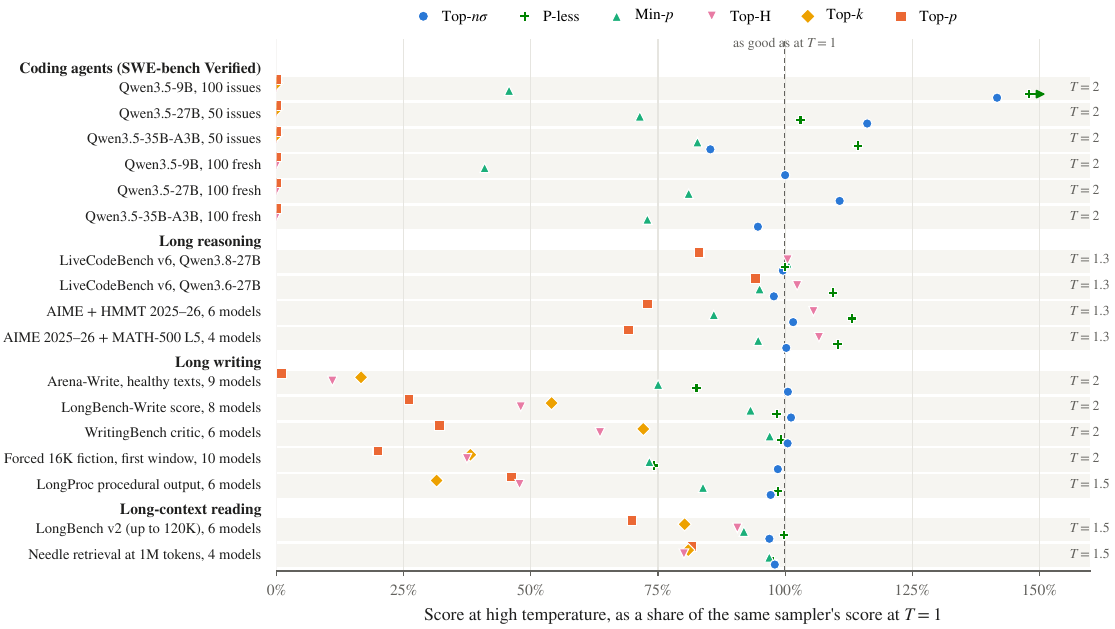}
\caption{ Each point is a sampler's score at the high temperature listed on the right, divided by that same sampler's score at temperature 1 on the same models and items, so 100\% (dashed line) means turning up the heat performed strongly. Dividing by each sampler's own baseline isolates how it reacts to temperature. Top-$n\sigma$ (blue circles) sits on the dashed line in every row, while top-$p$, top-$k$ and Top-H drift left and hit 0\% on coding agents. Arrows mark values above 150\% (P-less on Qwen3.5-9B resolves 50\% at temperature 2 and 33\% at 1). Reasoning rows use top-$p$ 0.95. Complete results are in Sections~\ref{sec:agentic}--\ref{sec:read} and the appendices.}
\label{fig:scoreboard}
\end{figure}

\subsection{Coding agents}
\label{sec:agentic}

\begin{figure}[t]
\centering
\includegraphics[width=\textwidth]{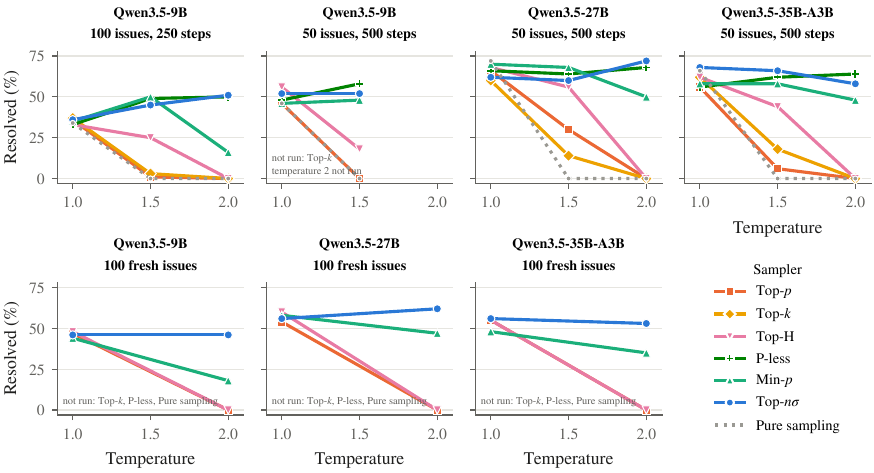}
\caption{SWE-bench Verified resolve rate against temperature for Qwen3.5 agents (mini-swe-agent, official test harness). The top row uses the original issue sets. The bottom row uses 100 fresh issues drawn with a fixed seed and no overlap with the first set. A line that stays flat means the agent keeps working as temperature rises. All samplers start together at temperature 1, the tail-admitting rules (orange, yellow, pink and dotted gray) fall to zero by temperature 2, and top-$n\sigma$, P-less and min-$p$ stay up. Absent lines and points were never run (the fresh-issue grids were pre-registered with four samplers, and the 500-step 9B grid only up to 1.5). The same split on an independently drawn issue set rules out a quirk of one sample. Gemma-4 is in Appendix~\ref{app:extra}.}
\label{fig:swe}
\end{figure}

The SWE-bench agent runs tens to a few hundred commands and edits per episode, its context grows to tens of thousands of tokens, and one malformed command can end the episode. Figure~\ref{fig:swe} has SWE-bench Verified resolve rates for seven Qwen3.5 configurations. Table~\ref{tab:swe} compares each sampler with top-$p$ issue by issue in our long-horizon setting (500 agent steps, 262K-token context).

\begin{table}[t]
\centering
\caption{Long-horizon coding agents (Qwen3.5-9B, 500-step limit, 262K context, 50 issues), the setting closest to how agents are run in practice. $T$ is the temperature. W/L counts issues the sampler resolved and top-$p$ 0.9 did not at the same temperature, and the reverse. A lopsided record like 26/0 means the same issues flip in one direction, which is very unlikely to be noise. $^{*}$ marks a two-sided sign test with $p<0.05$, and the min-$p$, P-less and top-$n\sigma$ wins at 1.5 all reach $p<10^{-6}$. Nothing is significant at 1.0, meaning samplers only differ once the temperature goes up.}
\label{tab:swe}
\small
\begin{tabular}{lcccc}
\toprule
 & \multicolumn{2}{c}{$T{=}1.0$} & \multicolumn{2}{c}{$T{=}1.5$} \\
\cmidrule(lr){2-3} \cmidrule(lr){4-5}
Sampler & Resolved & W/L vs top-$p$ & Resolved & W/L vs top-$p$ \\
\midrule
Pure sampling & 46\% & 6/6 & 0\% & 0/0 \\
Top-$p$ 0.9 & 46\% & base & 0\% & base \\
Min-$p$ 0.1 & 46\% & 7/7 & 48\% & 24/0$^{*}$ \\
Top-H 0.4 & 56\% & 9/4 & 18\% & 9/0$^{*}$ \\
\pless{} & 48\% & 6/5 & 58\% & 29/0$^{*}$ \\
Top-$n\sigma$ 1 & 52\% & 9/6 & 52\% & 26/0$^{*}$ \\
\bottomrule
\end{tabular}

\end{table}

There's not much to choose between the rules at temperature 1. They resolve 33--37\% of the 100-issue grid and 46--56\% in the long-horizon setting. Once the temperature goes up they split apart. On Qwen3.5-9B at temperature 2, top-$p$, top-$k$, Top-H and pure sampling resolve 0 of 100 issues, since their commands turn into garbled text within a few turns. Top-$n\sigma$ resolves 51 and P-less 50, and neither loses a single disputed issue (51--0 and 50--0). In the long-horizon setting the split already shows at 1.5. Min-$p$ gets 48\%, P-less 58\% and top-$n\sigma$ 52\%, top-$p$ gets 0\%, and the win--loss records are 24--0, 29--0 and 26--0.

 Qwen3.5-27B and Qwen3.5-35B-A3B don't resolve anything at temperature 2 under top-$p$, top-$k$, Top-H or pure sampling, and under top-$n\sigma$ they resolve 72\% and 58\% (36--0 and 29--0). We then drew 100 fresh issues with no overlap with the first set. Top-$n\sigma$ resolves 46, 62 and 53 percent of them on the 9B, 27B and 35B-A3B models while top-$p$ and Top-H resolve none, for records of 46--0, 62--0 and 53--0. An independent rerun of the long-horizon grid gives top-$n\sigma$ 56\% at 1.5 and 60\% at 2, compared with 2\% and 0\% for top-$p$.

Long chain-of-thought coding without an agent shows it too. On LiveCodeBench v6 with Qwen3.8-27B in thinking mode (175 problems, two samples, 32K-token budget, Table~\ref{tab:lcb}), every rule scores 61--64\% pass@1 up to temperature 1. At 1.3 top-$p$ 0.95 drops to 51\% while min-$p$, P-less, top-$n\sigma$ and Top-H stay at 63\% and win 49--53 disputed problem-samples for 6--8 losses. Qwen3.6-27B behaves the same way (Table~\ref{tab:lcb36}).

\subsection{Long-form writing, structured output and long-context reading}
\label{sec:bench}

\begin{figure}[t]
\centering
\includegraphics[width=\textwidth]{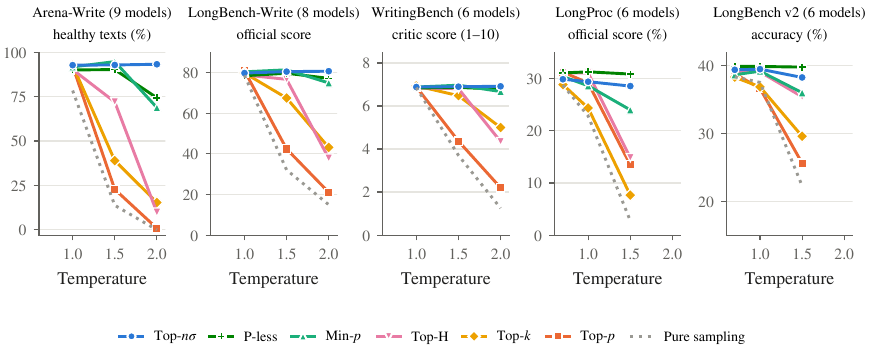}
\caption{From left to right, share of healthy Arena-Write generations (nine models, among them Llama-3.3-70B, gpt-oss-120b and Qwen3.5-122B-A10B), official LongBench-Write score, WritingBench critic score (555 queries), official LongProc score (six procedural tasks, outputs up to 8K tokens) and LongBench v2 accuracy (503 questions, contexts up to 120K tokens). The lines bunch together at temperature 1 and fan out as it rises, in the same order every time, with the tight rules on top. LongProc is graded by fixed rules and LongBench v2 by exact-match answers, so the ordering does not depend on an LLM judge. LongProc and LongBench v2 were run up to temperature 1.5. Tables~\ref{tab:bench}, \ref{tab:longproc} and~\ref{tab:lb2} have the numbers and paired tests.}
\label{fig:bench}
\end{figure}

Figure~\ref{fig:bench} covers Arena-Write~\citep{wu2026longwriterzero}, LongBench-Write~\citep{bai2025longwriter} and WritingBench~\citep{wu2025writingbench}, all with natural stopping. The rules sit close together at temperature 1 (90--93\% healthy on Arena-Write, 78--81 on LongBench-Write, 6.8--6.9 on WritingBench). At temperature 2 top-$n\sigma$ hasn't moved on any of the three (0.93 healthy, 80.6 and 6.9). P-less and min-$p$ lose some ground (0.74 and 0.69 healthy, 77 and 75 on LongBench-Write). Top-H, top-$k$ and top-$p$ break down badly. Only 0.10, 0.15 and 0.01 of their Arena-Write generations are healthy, they score 38, 43 and 21 on LongBench-Write, and top-$p$ gets 2.2 on WritingBench.

LongProc~\citep{ye2025longproc} asks for long \emph{structured} output, e.g. turning an HTML page into a TSV table or writing a travel plan, and grades it with fixed rules, so there's no judge in the loop. At 1.5 P-less and top-$n\sigma$ keep the 31\% and 29\% they had at temperature 1. Top-$p$ falls to 13\% and loses 532--73 and 469--74 on disputed items, top-$k$ falls to 8\% and pure sampling to 3\%. LongBench v2~\citep{bai2025longbenchv2} has the model read up to 120K tokens and reason before answering. Here P-less and top-$n\sigma$ hold 39.8\% and 38.3\% at 1.5, and top-$p$ drops from 36.6\% to 25.6\% (667--239 and 635--253). LongBench v2 is also the one benchmark where the tight rules lead even at temperature 1 (39.5\% for top-$n\sigma$ vs.\ 36.6\%, 355--269).

\subsection{Needle retrieval and long reasoning}
\label{sec:read}

Temperature hurts even when the model mostly just has to read. We hid 16 key--value ``needles'' in PG-19 novels between 8K and one million tokens long and asked for each value under every sampler (Figure~\ref{fig:niah} in Appendix~\ref{app:longctx}). At temperature 1 the rules retrieve about the same share at every length. At 1.5 the tight rules don't change, but every tail-admitting rule gets worse, and the longer the context the bigger the drop. At one million tokens top-$n\sigma$, P-less and min-$p$ retrieve 75\%, 78\% and 73\%, top-$p$ retrieves 59\% and pure sampling 48\% (106--25, 113--19 and 100--30 on queries where they disagree with top-$p$). Our explanation is that a longer context makes the model less sure of the answer. Under greedy decoding the entropy of the first answer token goes from 0.02 nats at 8K to 0.37 at one million tokens, and a rule that lets the tail in turns that extra uncertainty into wrong digits.

Long chain-of-thought math follows the same pattern (Appendix~\ref{app:reason}). At 1.3 top-$n\sigma$ matches its temperature-1 accuracy on AIME and HMMT 2025--26 with six models (40.2\% vs.\ 39.6\%) and on AIME plus MATH-500 level 5 with four models (57.9\% vs.\ 58.2\%). Top-$p$ 0.95 falls to 27.4\% and 39.2\%. Top-$n\sigma$ wins 219--29 and 326--36 on disputed problem-samples, and min-$p$, P-less and Top-H beat top-$p$ at 1.3 as well.

\subsection{Forced 16K-token generation}
\label{sec:r1}

\begin{figure}[t]
\centering
\includegraphics[width=\textwidth]{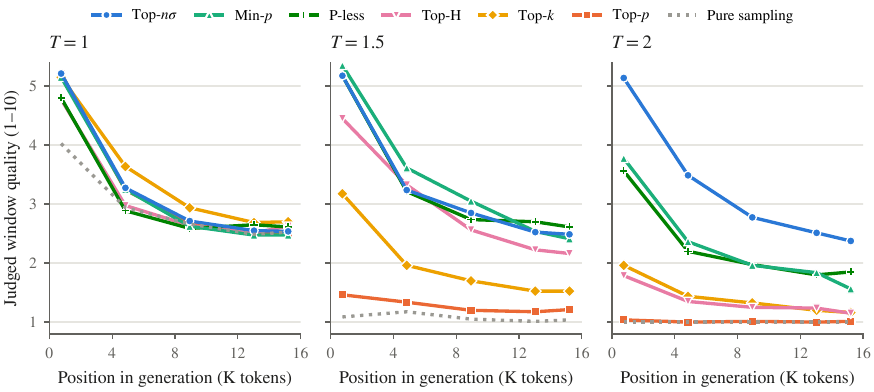}
\caption{Judged quality of 1.5K-token windows by position in forced 16K-token fiction, ten models from 1B to 120B, 80 generations per line. The left end of a line is how well the text opens and its slope is how fast it decays. At temperature 1 every rule opens near 5 and decays to about 2.5, and the sampler barely matters. At 1.5 top-$k$ shows the slow version of the failure, opening at 3.2 and sliding to 1.5. At 2 the tail-admitting rules are at the floor from the first window, i.e. word salad, while top-$n\sigma$ traces essentially the same curve it traces at temperature 1.}
\label{fig:hero}
\end{figure}

To see where along a long text things go wrong, we forced every model to write 16K tokens and judged it window by window (Figure~\ref{fig:hero}, Table~\ref{tab:r1}). At temperature 1 every truncation rule opens at 4.8--5.2 out of 10 and drifts down to about 2.5, so the sampler hardly matters for judged quality there. At temperature 2 it matters a great deal. Top-$p$, top-$k$, Top-H and pure sampling open at 1.0--2.0, which in practice means word salad from the first paragraph. Min-$p$ and P-less open at 3.8 and 3.6. Top-$n\sigma$ opens at 5.1, against 5.2 at temperature 1, and it is the only rule whose temperature-2 curve lies on top of its temperature-1 curve (last window 2.4 vs.\ 2.5).

\subsection{What and how many tokens are these samplers choosing?}
\label{sec:mechanism}

\begin{figure}[t]
\centering
\begin{minipage}[t]{0.48\textwidth}
\centering
\includegraphics[width=\linewidth]{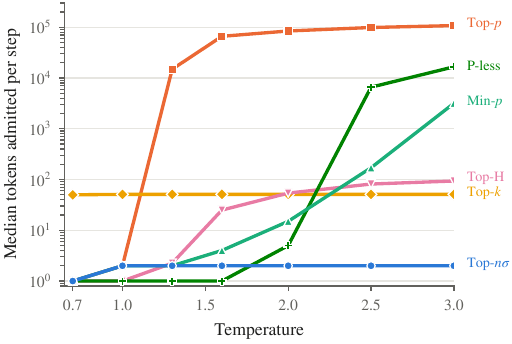}
\caption{Median tokens admitted per decoding step (log scale) against temperature, eight models, 64 generations per point. A flat line means temperature cannot pull new tokens into the set. Top-$n\sigma$ is flat at 1--2 tokens by construction and top-$k$ is flat at about 50, while other rules climb, top-$p$ fastest.}
\label{fig:pool}
\end{minipage}\hfill
\begin{minipage}[t]{0.48\textwidth}
\centering
\includegraphics[width=\linewidth]{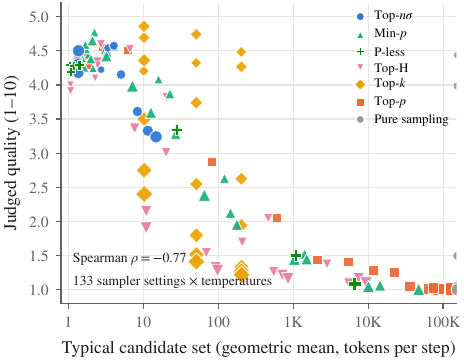}
\caption{Judged quality against the geometric-mean number of tokens admitted per step. Each point is one sampler setting at one temperature, averaged over eight models, and bigger markers mean higher temperature. Points from different rules fall on one downward curve (Spearman $\rho{=}{-}0.77$), The number of tokens a setting lets through predicts its quality without knowing which rule it is. 133 cells from 19 of the 22 settings in Figure~\ref{fig:map} (top-$n\sigma$ with $n$ of 1, 2 and 3 is missing because of a tag mismatch in our analysis script).}
\label{fig:predict}
\end{minipage}
\end{figure}

Our logits processor logs how many tokens each rule admits at every step, reported in Figure~\ref{fig:pool} and Table~\ref{tab:pool}. Top-$n\sigma$'s set can't change with temperature by construction, so its measured median is 1 in the ten-model generations and 2 in the eight-model sweep at every temperature. From temperature 1 to 2 the median grows from 1 to 14 tokens for min-$p$, to 21 for P-less and to 54 for Top-H, while top-$p$ ends up keeping most of the vocabulary. Means run higher than medians (min-$p$ and P-less sometimes admit thousands of tokens at temperature 2), and past 2.5 their sets get bigger than Top-H's. The faster a set grows, the more junk tokens temperature can promote. Those junk tokens are what turn a story into word salad or a shell command into a format error. We think of long-context failure as an accumulation of these small sampling errors. Each one pushes the model a bit further out of distribution, until it can't find its way back. If one also accounts for how big the set is at temperature 1 (top-$k$ never grows, but it always admits about 50 tokens), this measurement sorts the rules into the same groups the benchmarks do.

\subsection{Sweeping hyperparameters}
\label{sec:map}

\begin{figure}[t]
\centering
\includegraphics[width=\textwidth]{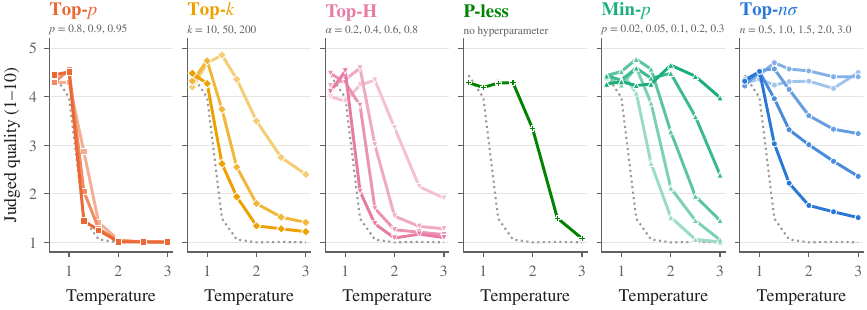}
\caption{Mean judged window quality against temperature for all 22 settings in our sweep, eight models from five families, 64 forced 8K-token generations per point. Each panel is one rule. Lighter to darker lines follow the listed hyperparameter values, and dotted gray is pure sampling for reference. If the cliff were a tuning problem, every rule would have some setting that stays flat. Only top-$n\sigma$ with $n\le1$ does, and min-$p$ with $p$ of 0.3 nearly does. Elsewhere the hyperparameter only moves the cliff, and P-less (no hyperparameter) holds until 1.6 and then falls.}
\label{fig:map}
\end{figure}

An obvious objection is that all of this comes down to hyperparameter choice. To check, we swept 22 settings over seven temperatures from 0.7 to 3.0 on eight models from five families, with 64 forced 8K-token generations per cell (Figure~\ref{fig:map}). Top-$n\sigma$ with $n$ of 1 or less, whose set is small, scores 4.2--4.7 from 0.7 all the way up to 3.0. Larger $n$ lets in more of the tail at every temperature, and the cliff comes back. For the other families the hyperparameter just moves the cliff around. Every top-$p$ setting is below 1.5 by temperature 1.6. Min-$p$ lasts longer as $p$ goes up ($p$ of 0.02 collapses after 1.3, 0.1 after 1.6, and 0.3 is still above 4 at 2.5), though you pay for that with a tighter set at all the lower temperatures. If we pool the 133 setting-by-temperature cells that have per-step candidate-set logs (19 of the 22 settings, see Figure~\ref{fig:predict}), the typical number of tokens a setting admits per step predicts its judged quality with a Spearman correlation of $-0.77$, whichever rule produced it. Nothing stays usable above roughly 100 tokens per step.

\begin{figure}[t]
\centering
\includegraphics[width=\textwidth]{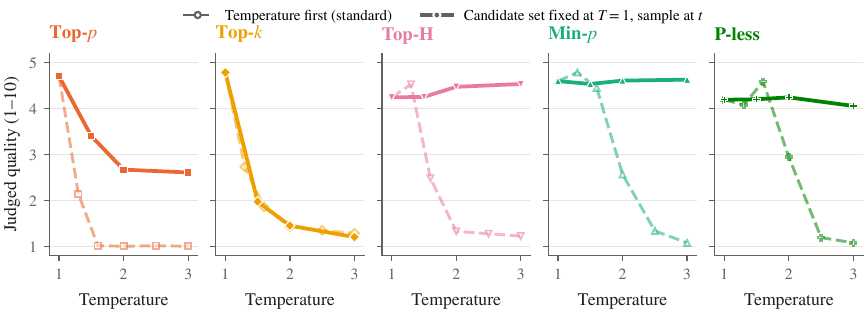}
\caption{The causal test. Dashed lines apply temperature first, as inference engines do by default. Solid lines choose the candidate set at temperature 1 and only then sample at the temperature on the x-axis, so the set can't grow. Four models, 32 generations per point. If flatter probabilities inside a good set were the problem, solid lines would fall with dashed ones, and for Top-H, min-$p$ and P-less they don't fall at all. Top-$p$ and top-$k$ are helped only partly or not at all, because their temperature-1 sets already reach into the tail.}
\label{fig:tl}
\end{figure}

\subsection{Applying temperature last}
\label{sec:ablation}

If high temperature hurts because it changes \emph{which} tokens get in, then picking the candidate set at temperature 1 and only applying temperature inside it should make the cliff go away. With the set picked at temperature 1, Top-H, min-$p$ and P-less keep their temperature-1 quality all the way to 3 (4.1--4.6), where the default order has them down at 1.1--1.2. This only rescues rules whose temperature-1 set already leaves out the tail. Top-$p$ at temperature 1 admits a mean of 73 tokens in the ten-model generations, because on uncertain steps it reaches deep into the tail anyway, and top-$k$ always admits about 50. Fixing their sets helps them partly or not at all. The result also holds on LongProc with its rule-based scoring (Appendix~\ref{app:longctx}). With the set picked at temperature 1, min-$p$ 0.1 scores 25\% at temperature 2 instead of 15\%, and top-$p$ 0.95 comes back from 1\% to 25\% at 1.5 (Qwen3.5-4B and 9B, five tasks). Leaving out the tail is what matters, and top-$n\sigma$ does it at every temperature by design.

\subsection{Robustness}
\label{sec:robust}

The forced-length and benchmark results survive the checks in Appendix~\ref{app:robust}, which covers a second seed on all ten core models, six newer models up to Qwen3.5-122B-A10B (top-$n\sigma$ opens at 7.4 at temperature 2 and top-$p$ at 1.7), natural-length generation where the model may stop and is asked to continue, and non-fiction. The second judge from another family (GLM-4.5-Air) agrees with the first at a Spearman correlation of 0.92 on cell means, The second seed reproduces the 220 model-by-sampler-by-temperature cells at 0.95 (Figure~\ref{fig:agree}).

\section{Practical Guidance}
\label{sec:discussion}

\paragraph{If you turn up the temperature, use a sampler that leaves out the tail.}
A rule whose candidate set grows with temperature ties together two knobs, i.e. how much randomness to add among reasonable tokens and how far to reach into unreasonable ones. Top-$n\sigma$ with $n$ of 1 or less keeps them apart by construction, and it is already available in llama.cpp. Min-$p$ with a larger $p$, or any tight rule whose set is picked at temperature 1, keeps them apart in practice.

\paragraph{Report what the sampler keeps!}
Benchmarks run only at temperature 1 hide most of these differences, and nominal hyperparameters hide the property that matters, so sampler comparisons should report the realized candidate-set size per step. Benchmark results for open-weight models should also list the full sampler configuration and its order, since the same model at the same temperature goes from 62\% of issues resolved to 0\% on the sampler alone.

\section{Conclusion}
\label{sec:conclusion}

The sampler is a free and extremely effective control over long-context generation, Yet hardly anyone tunes it or even reports it. Across 29 models and every benchmark tried, samplers whose candidate set stays small and stable lose little or nothing at high temperature, and samplers that admit the tail fall apart. Long-context generation is, to a large degree, a sampling problem once you turn up the heat.

\begin{ack}
This work was funded by Thoughtworks.
\end{ack}

\bibliography{refs}
\bibliographystyle{plainnat}

\newpage
\appendix

\section{Sampler definitions and temperature}
\label{app:defs}

Let $\mathbf{z}\in\mathbb{R}^{|V|}$ be the logits at one step, $T$ the temperature, $p=\mathrm{softmax}(\mathbf{z}/T)$ the tempered distribution, and $\mathcal{C}\subseteq V$ the candidate set. Following HuggingFace, vLLM and SGLang, every rule is applied to $p$, i.e. after temperature, except in the temperature-last experiments of Section~\ref{sec:ablation}.

\paragraph{Top-$n\sigma$} keeps every token within $n$ standard deviations of the maximum logit:
\begin{equation}
\mathcal{C}_{\text{n-sigma}} = \{v \in V : z_v \ge \max_{u} z_u - n \cdot \sigma(\mathbf{z})\}.
\label{eq:nsigma}
\end{equation}
\paragraph{P-less} keeps tokens at least as probable as the collision probability of two independent draws, $\mathcal{C}_{\text{p-less}} = \{v : p(v) \ge \sum_u p(u)^2\}$. \textbf{Min-$p$} keeps $\{v : p(v) \ge p_{\min}\cdot\max_u p(u)\}$. \textbf{Top-H} adds tokens in descending probability until the entropy of the renormalized set reaches a fraction $\alpha$ of $H(p)$, with $\alpha{=}0.4$. \textbf{Top-$p$} keeps the smallest prefix of tokens sorted by probability whose total mass reaches $p$, and \textbf{top-$k$} keeps the $k$ most probable tokens.

\paragraph{Temperature invariance.}
Replacing $\mathbf{z}$ by $\mathbf{z}/T$ scales both $\max_u z_u$ and $\sigma(\mathbf{z})$ by $1/T$, so the condition in Equation~\ref{eq:nsigma} is unchanged and the top-$n\sigma$ set is identical at every $T{>}0$~\citep{tang2024topnsigma}. Top-$k$ is invariant by construction. Min-$p$, P-less, Top-H and top-$p$ all threshold on $p$, which flattens as $T$ grows, so their sets grow with $T$. The min-$p$ and P-less thresholds are tied to the head of $p$ ($\max_u p(u)$ and $\sum_u p(u)^2$), which shrinks slowly, while the top-$p$ and Top-H thresholds are tied to cumulative mass and to $H(p)$, which react much faster.

\paragraph{Temperature last.}
In Section~\ref{sec:ablation} the candidate set is computed from $\mathrm{softmax}(\mathbf{z})$, i.e. at $T{=}1$, and the token is then drawn from $\mathrm{softmax}(\mathbf{z}/t)$ restricted to that set. We write $t$ for the sampling temperature in this setting.

\section{Models}
\label{app:models}
The ten-model core set is Llama-3.2-1B, Qwen2.5-1.5B, Qwen2.5-3B, Mistral-7B-v0.3, Llama-3.1-8B, Mistral-Nemo-12B, Qwen2.5-14B, Qwen2.5-32B, Llama-3.3-70B~\citep{dubey2024llama3,yang2024qwen2,jiang2023mistral} and gpt-oss-120b~\citep{openai2025gptoss}. Other experiments add Qwen3.5 (9B, 27B, 35B-A3B, 122B-A10B), Qwen3.6-27B, Qwen3.8-27B, Gemma-4 (E4B, 12B, 31B), OLMo-3 (7B, 3.1-32B), Nemotron-3 (Nano-4B, Nano-30B-A3B) and GLM-4.5-Air, and the long-context and reasoning benchmarks also use Qwen3.5 (2B, 4B), gpt-oss-20b, DeepSeek-R1-Distill-Qwen-1.5B and the one-million-token Llama-3.1-8B-UltraLong~\citep{xu2025ultralong}. All are the instruction-tuned releases.

\section{Additional analyses}
\label{app:extra}

\subsection{Behavior at the default temperature and under forced length}
Forced out to 16K tokens at $T{=}1$, top-$p$, min-$p$, Top-H, \pless{} and top-$n\sigma$ end in a terminal loop 95--100\% of the time, which matches greedy decoding. Top-$k$ and pure sampling loop 56\% and 41\% of the time (Table~\ref{tab:r1}). No setting gets you a healthy long generation. Across the whole sweep no cell is healthy more than 31\% of the time, and top-$n\sigma$ with $n{=}1$ still ends in a terminal loop 44--69\% of the time at every temperature. So top-$n\sigma$ makes a much wider range of temperatures usable, but a model forced to 16K tokens still collapses. Higher temperature only gets rid of loops for the tail-admitting rules by turning the text into word salad, and top-$n\sigma$ still loops in 71\% of generations at $T{=}2$. What the sampler controls is how a forced long generation responds to temperature. A model that tends to loop will still loop.

\subsection{Repetition penalties and an escape floor}
\label{sec:stack}

If loops are the real problem and letting in the tail is the cost of getting out of them, two additions seem worth trying. One is a repetition penalty. The other is a floor that always keeps the top two tokens, so a tight rule can never go fully greedy. We put both on top of each tight and tail-admitting rule on eight of the ten models, Llama-3.3-70B and gpt-oss-120b included (Table~\ref{tab:stack}). DRY cuts the $T{=}1$ loop rate from 95--100\% to 27--38\% for every rule, but judged quality moves by 0.2 points at most in either direction at any temperature. The penalty swaps verbatim loops for text that is more varied and no better. The top-2 floor does almost nothing at $T{=}1$ for a simple reason, which is that a second candidate with probability $10^{-6}$ never gets picked. It only does something once temperature is raised. At $T{=}2$, top-$n\sigma$ with the floor nearly eliminates terminal loops (75\% down to 2\%) and keeps its judged quality (3.3). It is the only loop-free configuration at $T{=}2$ that isn't word salad, since top-$p$, Top-H, min-$p$ and \pless{} with the floor score 1.0, 1.3, 2.4 and 2.5. A tail-excluding rule plus an escape floor plus higher temperature gets rid of loops without letting in the noisy tail. It still doesn't produce good long text, because these models have a low quality ceiling at 16K under every sampler. On Arena-Write with natural stopping (six models), where loops are already rare, the floor changes little. Top-$n\sigma$ is healthy on 0.94 of generations at $T{=}2$ with the floor and 0.93 without, and the floor even lowers \pless{} at $T{=}2$ (0.60 vs.\ 0.70).

For coding agents the floor hurts. Adding it to top-$n\sigma$ drops agents to 42\% and 22\%, even though it gets rid of loops in fiction. The floor admits a second token at every step, including the steps where the agent has to type an exact command, and format errors per episode go from 0.04 to 1.4 at $T{=}1.5$ and from 0.06 to 9.6 at $T{=}2$.

\subsection{Context depth}
\label{sec:depth}

Given the title, it's fair to ask whether a longer \emph{input} context by itself pushes models into the regime where the sampler matters. We continued PG-19 novels~\citep{rae2019compressive} from depths of 2K, 16K and 64K tokens on six long-context models, and from 128K on the three that support it. We measured the model's entropy at $T{=}1$ and the judged-quality gap between tail-excluding rules (min-$p$, \pless{}, top-$n\sigma$) and tail-admitting ones (top-$p$, top-$k$, Top-H, pure sampling) (Table~\ref{tab:depth}). Our answer is no. The gap stays within 0.4 points of zero at $T{\le}1$ at every depth and sits at 1.2--1.6 points at $T{=}1.3$ and $1.6$ at every depth, with no trend. Entropy at $T{=}1$ rises with depth on Qwen3.5-27B and Qwen3.5-35B-A3B (from 1.16 to 1.70 nats on the latter), is mixed on Qwen3.5-9B and Nemotron, is near zero on Gemma-4-12B, and falls on Llama-3.1-8B. In our data the sampling problem comes from how much the model writes, since that is where loops form and where temperature gets used to escape them. How much it reads has no measurable effect on the gap.

\subsection{Other model families in agentic coding}
\label{app:gemma}
Gemma-4-31B is the main exception among coding agents, Its unique candidate sets explain this. No rule separates from the others at $T{=}1.5$, either on the first 50 instances or on the 100 fresh ones (49--55\% for every rule, all $p{>}0.3$). In its forced-length generations (Table~\ref{tab:r10}), top-$p$ still admits a median of one token per step at $T{=}1.5$ (33 at $T{=}2$), where Qwen3.5-27B, OLMo-3.1-32B and Llama-3.1-8B admit 28,000--143,000. We think Gemma's soft-capped final logits are responsible, since they keep the distribution peaked under temperature, but we didn't isolate the cap experimentally. Gemma-4-E4B resolves at most 12\% under any rule, too few to compare. We couldn't evaluate OLMo-3.1-32B at all. Under every rule, $T{=}1$ included, it emits several actions per turn, mini-swe-agent ends 95\% of its episodes (over 10 configurations) for repeated format errors, and it resolves no instances.

\begin{figure}[h]
\centering
\includegraphics[width=0.6\textwidth]{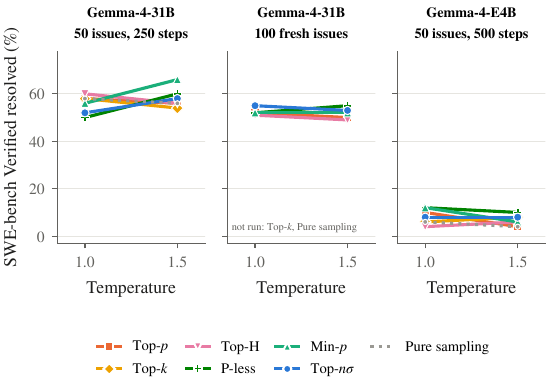}
\caption{SWE-bench Verified resolve rate for Gemma-4 agents. Unlike Qwen3.5, the samplers hardly separate at $T{=}1.5$ on either Gemma-4-31B issue set. Gemma-4-E4B resolves too few issues to compare. $T{=}2$ was not run on Gemma.}
\label{fig:swe_gemma}
\end{figure}

\section{Coding agents in detail}
\label{app:swe}

\begin{figure}[h]
\centering
\includegraphics[width=\textwidth]{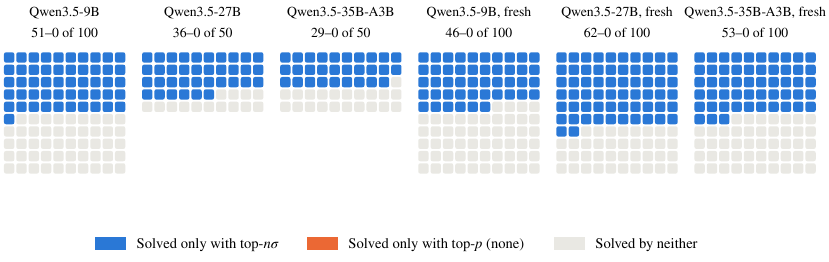}
\caption{SWE-bench Verified issues at $T{=}2$, one square each, sorted by outcome, for six model-by-issue-set grids. There are 277 blue squares and no orange ones, so the gap in Figure~\ref{fig:swe} is not an average over trade-offs. Top-$n\sigma$ wins every issue on which the two agents disagree.}
\label{fig:disputed}
\end{figure}

\begin{figure}[h]
\centering
\includegraphics[width=\textwidth]{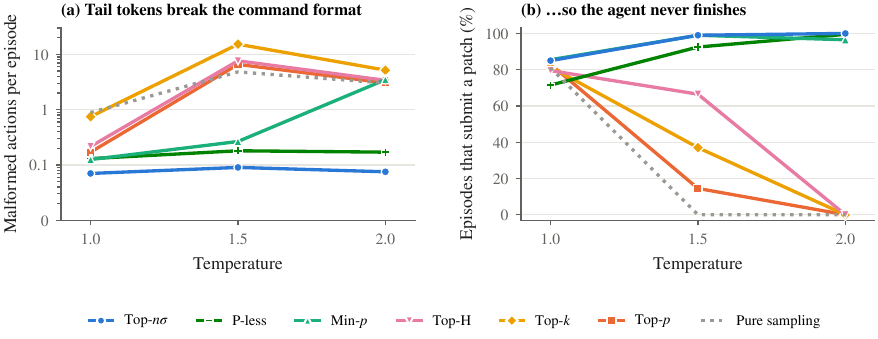}
\caption{Qwen3.5-9B, 27B and 35B-A3B agents on the original issue sets (4,200 episodes). (a) Malformed actions per episode (log scale), i.e. commands the harness can't parse. (b) Share of episodes that end with a submitted patch. Panel (a) is the cause and panel (b) the effect. At $T{=}2$ the top-$p$, top-$k$, Top-H and pure-sampling agents emit 3--5 malformed actions within their first few turns and never submit, while top-$n\sigma$ agents emit 0.07 per episode and submit every time. The failures happen at the level of command format, before the quality of the agent's reasoning even comes into play.}
\label{fig:swe_anatomy}
\end{figure}

Figure~\ref{fig:swe_anatomy} shows how the agents in Figure~\ref{fig:swe} fail. A shell command or code edit has to be exact, and one tail token, like a stray Unicode character or a broken quote, is enough to make it unparsable. At $T{=}1.5$ top-$k$ agents average 15 malformed actions per episode and top-$p$ agents 6.6, compared with 0.08 for top-$n\sigma$. At $T{=}2$ the tail-admitting agents get terminated after 3--9 steps. Tight rules keep the format intact at every temperature, so their agents spend their steps working on the task. At $T{=}1.5$ and $2$ they submit a patch in 92--100\% of episodes, more often than at $T{=}1$ (71--86\%), where some episodes run out of steps.

\subsection{SWE-bench Verified: 100-instance grid}

Qwen3.5-9B, mini-swe-agent, 250-step limit, 100 fixed instances, three temperatures. As is standard for SWE-bench, resolve rates are over all instances, and empty patches or patches the official harness couldn't evaluate count as unresolved. W/L pairs each rule with top-$p$ 0.9 on the same instances. At $T{=}1$ every rule resolves 33--37\% and no paired comparison is significant ($p{>}0.6$). At $T{=}1.5$ and $2$ the tail-excluding rules win every disagreement.

\begin{table}[h]
\centering
\caption{Qwen3.5-9B on 100 fixed SWE-bench Verified issues (250-step limit), W/L as in Table~\ref{tab:swe}. At $T{=}1$ every rule resolves 33--37\% and nothing is significant, and at $T{=}1.5$ and 2 every disagreement goes to the tail-excluding rules. This grid is the first group in Figure~\ref{fig:teaser}(b).}
\label{tab:swe100}
\small
\begin{tabular}{lcccccc}
\toprule
 & \multicolumn{2}{c}{$T{=}1.0$} & \multicolumn{2}{c}{$T{=}1.5$} & \multicolumn{2}{c}{$T{=}2.0$} \\
\cmidrule(lr){2-3} \cmidrule(lr){4-5} \cmidrule(lr){6-7}
Sampler & Resolved & W/L vs top-$p$ & Resolved & W/L vs top-$p$ & Resolved & W/L vs top-$p$ \\
\midrule
Pure sampling & 34\% & 7/9 & 0\% & 0/1 & 0\% & 0/0 \\
Top-$k$ 50 & 37\% & 10/9 & 3\% & 3/1 & 0\% & 0/0 \\
Top-$p$ 0.9 & 36\% & base & 1\% & base & 0\% & base \\
Min-$p$ 0.1 & 35\% & 12/13 & 50\% & 49/0$^{*}$ & 16\% & 16/0$^{*}$ \\
Top-H 0.4 & 33\% & 10/13 & 25\% & 24/0$^{*}$ & 0\% & 0/0 \\
\pless{} & 33\% & 8/11 & 49\% & 48/0$^{*}$ & 50\% & 50/0$^{*}$ \\
Top-$n\sigma$ 1 & 36\% & 9/9 & 45\% & 44/0$^{*}$ & 51\% & 51/0$^{*}$ \\
\bottomrule
\end{tabular}

\end{table}

\subsection{Larger models, fresh issues and an independent repeat}

\begin{table}[h]
\centering
\caption{Qwen3.5-27B (top) and Qwen3.5-35B-A3B (bottom), 50 issues, 500-step limit, 262K context. W/L is against top-$p$ 0.9 at the same $T$, and $^{*}$ marks a sign test with $p<0.05$. One dense and one mixture-of-experts model show the same split as the 9B model in Table~\ref{tab:swe}, so the effect is not specific to one model size or architecture.}
\label{tab:swe_r7}
\small
\begin{tabular}{lcccccc}
\toprule
 & \multicolumn{2}{c}{$T{=}1.0$} & \multicolumn{2}{c}{$T{=}1.5$} & \multicolumn{2}{c}{$T{=}2.0$} \\
\cmidrule(lr){2-3} \cmidrule(lr){4-5} \cmidrule(lr){6-7}
Sampler & Resolved & W/L vs top-$p$ & Resolved & W/L vs top-$p$ & Resolved & W/L vs top-$p$ \\
\midrule
Pure sampling & 72\% & 6/3 & 0\% & 0/15$^{*}$ & 0\% & 0/0 \\
Top-$k$ 50 & 60\% & 1/4 & 14\% & 0/8$^{*}$ & 0\% & 0/0 \\
Top-$p$ 0.9 & 66\% & base & 30\% & base & 0\% & base \\
Min-$p$ 0.1 & 70\% & 5/3 & 68\% & 20/1$^{*}$ & 50\% & 25/0$^{*}$ \\
Top-H 0.4 & 68\% & 5/4 & 56\% & 15/2$^{*}$ & 0\% & 0/0 \\
\pless{} & 66\% & 3/3 & 64\% & 17/0$^{*}$ & 68\% & 34/0$^{*}$ \\
Top-$n\sigma$ 1 & 62\% & 3/5 & 60\% & 15/0$^{*}$ & 72\% & 36/0$^{*}$ \\
\bottomrule
\end{tabular}
\\[4pt]
\begin{tabular}{lcccccc}
\toprule
 & \multicolumn{2}{c}{$T{=}1.0$} & \multicolumn{2}{c}{$T{=}1.5$} & \multicolumn{2}{c}{$T{=}2.0$} \\
\cmidrule(lr){2-3} \cmidrule(lr){4-5} \cmidrule(lr){6-7}
Sampler & Resolved & W/L vs top-$p$ & Resolved & W/L vs top-$p$ & Resolved & W/L vs top-$p$ \\
\midrule
Pure sampling & 66\% & 8/3 & 0\% & 0/3 & 0\% & 0/0 \\
Top-$k$ 50 & 62\% & 6/3 & 18\% & 7/1 & 0\% & 0/0 \\
Top-$p$ 0.9 & 56\% & base & 6\% & base & 0\% & base \\
Min-$p$ 0.1 & 58\% & 5/4 & 58\% & 26/0$^{*}$ & 48\% & 24/0$^{*}$ \\
Top-H 0.4 & 62\% & 7/4 & 44\% & 20/1$^{*}$ & 0\% & 0/0 \\
\pless{} & 56\% & 5/5 & 62\% & 28/0$^{*}$ & 64\% & 32/0$^{*}$ \\
Top-$n\sigma$ 1 & 68\% & 9/3 & 66\% & 30/0$^{*}$ & 58\% & 29/0$^{*}$ \\
\bottomrule
\end{tabular}

\end{table}

\begin{table}[h]
\centering
\caption{100 fresh SWE-bench Verified issues with no overlap with the first 100 (seed 20261101), 250-step limit. From top to bottom, Qwen3.5-9B, 27B and 35B-A3B. These issues were drawn and pre-registered after the first grids were analyzed, which makes them an out-of-sample check, and top-$n\sigma$ wins them 46--0, 62--0 and 53--0 at $T{=}2$. Only the four samplers listed were run.}
\label{tab:swe_fresh}
\small
\begin{tabular}{lcccc}
\toprule
 & \multicolumn{2}{c}{$T{=}1.0$} & \multicolumn{2}{c}{$T{=}2.0$} \\
\cmidrule(lr){2-3} \cmidrule(lr){4-5}
Sampler & Resolved & W/L vs top-$p$ & Resolved & W/L vs top-$p$ \\
\midrule
Top-$p$ 0.9 & 47\% & base & 0\% & base \\
Min-$p$ 0.1 & 44\% & 13/16 & 18\% & 18/0$^{*}$ \\
Top-H 0.4 & 48\% & 9/8 & 0\% & 0/0 \\
Top-$n\sigma$ 1 & 46\% & 10/11 & 46\% & 46/0$^{*}$ \\
\bottomrule
\end{tabular}
\\[4pt]
\begin{tabular}{lcccc}
\toprule
 & \multicolumn{2}{c}{$T{=}1.0$} & \multicolumn{2}{c}{$T{=}2.0$} \\
\cmidrule(lr){2-3} \cmidrule(lr){4-5}
Sampler & Resolved & W/L vs top-$p$ & Resolved & W/L vs top-$p$ \\
\midrule
Top-$p$ 0.9 & 54\% & base & 0\% & base \\
Min-$p$ 0.1 & 58\% & 7/3 & 47\% & 47/0$^{*}$ \\
Top-H 0.4 & 60\% & 8/2 & 0\% & 0/0 \\
Top-$n\sigma$ 1 & 56\% & 6/4 & 62\% & 62/0$^{*}$ \\
\bottomrule
\end{tabular}
\\[4pt]
\begin{tabular}{lcccc}
\toprule
 & \multicolumn{2}{c}{$T{=}1.0$} & \multicolumn{2}{c}{$T{=}2.0$} \\
\cmidrule(lr){2-3} \cmidrule(lr){4-5}
Sampler & Resolved & W/L vs top-$p$ & Resolved & W/L vs top-$p$ \\
\midrule
Top-$p$ 0.9 & 55\% & base & 0\% & base \\
Min-$p$ 0.1 & 48\% & 6/13 & 35\% & 35/0$^{*}$ \\
Top-H 0.4 & 55\% & 10/10 & 0\% & 0/0 \\
Top-$n\sigma$ 1 & 56\% & 6/5 & 53\% & 53/0$^{*}$ \\
\bottomrule
\end{tabular}

\end{table}

\begin{table}[h]
\centering
\caption{Independent repeat of the long-horizon grid in Table~\ref{tab:swe} (Qwen3.5-9B, 50 issues, 500 steps). Top-$n\sigma$ gets 56\% and 60\% at $T{=}1.5$ and $2$ against 2\% and 0\% for top-$p$, so Table~\ref{tab:swe} is not a lucky run. Dashes are cells that were not run.}
\label{tab:swe_s2}
\small
\begin{tabular}{lcccccc}
\toprule
 & \multicolumn{2}{c}{$T{=}1.0$} & \multicolumn{2}{c}{$T{=}1.5$} & \multicolumn{2}{c}{$T{=}2.0$} \\
\cmidrule(lr){2-3} \cmidrule(lr){4-5} \cmidrule(lr){6-7}
Sampler & Resolved & W/L vs top-$p$ & Resolved & W/L vs top-$p$ & Resolved & W/L vs top-$p$ \\
\midrule
Pure sampling & 30\% & 5/9 & 0\% & 0/1 & -- & -- \\
Top-$k$ 50 & 30\% & 3/7 & 2\% & 1/1 & -- & -- \\
Top-$p$ 0.9 & 38\% & base & 2\% & base & 0\% & base \\
Min-$p$ 0.1 & 36\% & 5/6 & 42\% & 20/0$^{*}$ & -- & -- \\
Top-H 0.4 & 36\% & 3/4 & 24\% & 12/1$^{*}$ & -- & -- \\
\pless{} & 30\% & 3/7 & 46\% & 22/0$^{*}$ & -- & -- \\
Top-$n\sigma$ 1 & 42\% & 7/5 & 56\% & 27/0$^{*}$ & 60\% & 30/0$^{*}$ \\
\bottomrule
\end{tabular}

\end{table}

\section{Long reasoning}
\label{app:reason}

\begin{figure}[h]
\centering
\includegraphics[width=\textwidth]{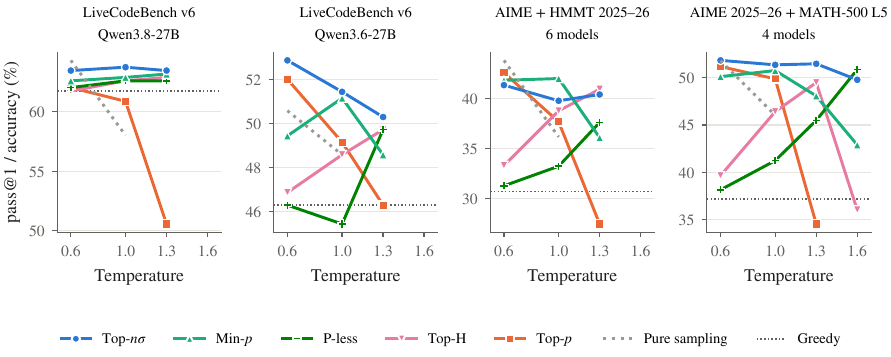}
\caption{Long chain-of-thought reasoning in thinking mode, 32K-token budget, two samples per problem. The panels are LiveCodeBench v6 on two models (175 problems, sandboxed official tests), AIME and HMMT 2025--26 from MathArena (123 problems on Qwen3.5-2B, 4B, 9B, gpt-oss-20b, Nemotron-3-Nano-30B-A3B and DeepSeek-R1-Distill-Qwen-1.5B), and AIME 2025--26 plus MATH-500 level 5 (194 problems on Qwen3.5-2B, 4B, Nemotron-3-Nano-4B and R1-Distill-Qwen-1.5B). Reasoning grids use top-$p$ 0.95, and dotted lines are greedy decoding. A reasoning trace runs to tens of thousands of tokens, so this is long generation too. Top-$n\sigma$ is flat from $T{=}0.6$ to $1.3$ while top-$p$ falls, and on the math benchmarks greedy decoding stays well below top-$n\sigma$ at every temperature because its traces loop.}
\label{fig:reason}
\end{figure}

Thinking models write tens of thousands of tokens before they answer, so reasoning traces are long generations in their own right. Figure~\ref{fig:reason} and Tables~\ref{tab:reason2}--\ref{tab:reason} show the pattern from the main text, only at a lower temperature scale. At $T{=}1.3$ top-$p$ 0.95 loses 10 to 18 points on the math benchmarks and 10 on LiveCodeBench with Qwen3.8-27B. Top-$n\sigma$ loses about one point at most and beats top-$p$ in every grid at $T{=}1.3$. Greedy decoding turns out to be the worst option on both math benchmarks (30.6\% on AIME and HMMT, 42.1\% on AIME and MATH-500 level 5), because 50--61\% of its traces end in a reasoning loop. Top-$n\sigma$ beats it by 9 to 16 points at every temperature, with 3--21\% of its traces looping.

\begin{table}[h]
\centering
\caption{AIME and HMMT 2025--26, six models, two samples per problem (pass@1, \%). W/L counts problem-samples the rule solved and top-$p$ 0.95 didn't at the same $T$, and the reverse. $^{*}$ marks a sign test with $p<0.05$. At $T{=}1.3$ top-$n\sigma$ stays within a point of its $T{=}1$ accuracy while top-$p$ drops by more than ten.}
\label{tab:reason2}
\small
\begin{tabular}{lrrrrrr}
\toprule
& \multicolumn{2}{c}{$T{=}0.6$} & \multicolumn{2}{c}{$T{=}1$} & \multicolumn{2}{c}{$T{=}1.3$} \\
\cmidrule(lr){2-3} \cmidrule(lr){4-5} \cmidrule(lr){6-7}
Sampler & Score & W/L & Score & W/L & Score & W/L \\
\midrule
Top-$p$ 0.95 & 42.4 &  & 37.5 &  & 27.4 &  \\
Top-H 0.4 & 33.2 & 50--186$^{*}$ & 38.6 & 127--111 & 40.8 & 228--30$^{*}$ \\
Min-$p$ 0.1 & 41.7 & 102--113 & 41.8 & 137--74$^{*}$ & 36.0 & 157--30$^{*}$ \\
P-less & 31.2 & 59--225$^{*}$ & 33.1 & 101--166$^{*}$ & 37.5 & 216--67$^{*}$ \\
Top-$n\sigma$ 1 & 41.2 & 100--118 & 39.6 & 118--87$^{*}$ & 40.2 & 219--29$^{*}$ \\
\bottomrule
\end{tabular}
\end{table}

\begin{table}[h]
\centering
\caption{AIME 2025--26 and MATH-500 level 5, four models, two samples per problem. Columns as in Table~\ref{tab:reason2}. The pattern matches Table~\ref{tab:reason2} on a different problem set and mostly different models.}
\label{tab:reason}
\small
\begin{tabular}{lrrrrrr}
\toprule
& \multicolumn{2}{c}{$T{=}0.6$} & \multicolumn{2}{c}{$T{=}1$} & \multicolumn{2}{c}{$T{=}1.3$} \\
\cmidrule(lr){2-3} \cmidrule(lr){4-5} \cmidrule(lr){6-7}
Sampler & Score & W/L & Score & W/L & Score & W/L \\
\midrule
Top-$p$ 0.95 & 57.2 &  & 57.1 &  & 39.2 &  \\
Top-H 0.4 & 44.6 & 71--267$^{*}$ & 51.8 & 89--171$^{*}$ & 55.3 & 324--73$^{*}$ \\
Min-$p$ 0.1 & 55.3 & 110--139 & 57.7 & 113--104 & 54.2 & 287--54$^{*}$ \\
P-less & 43.6 & 63--274$^{*}$ & 46.4 & 78--244$^{*}$ & 51.2 & 286--100$^{*}$ \\
Top-$n\sigma$ 1 & 57.4 & 125--122 & 58.2 & 116--99 & 57.9 & 326--36$^{*}$ \\
\bottomrule
\end{tabular}
\end{table}

\section{Long procedural output and long-context reading}
\label{app:longctx}

\begin{figure}[h]
\centering
\includegraphics[width=\textwidth]{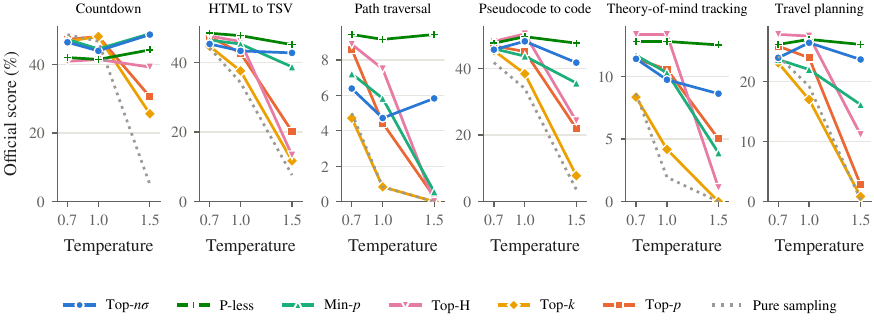}
\caption{LongProc broken out by task. Official rule-based scores for six models (Qwen3.5-4B, 9B, 27B, Gemma-4-E4B, Llama-3.1-8B, Nemotron-3-Nano-30B-A3B), 60 items per task. P-less and top-$n\sigma$ are flat or nearly flat on every task, and top-$k$ and pure sampling collapse on every task at $T{=}1.5$, so the average in Figure~\ref{fig:bench} is not carried by one task.}
\label{fig:longproc}
\end{figure}

\begin{table}[h]
\centering
\caption{LongProc, mean official score over six tasks and six models (\%). W/L counts items the rule scored higher on than top-$p$ 0.9 at the same $T$, and the reverse. The P-less and top-$n\sigma$ wins over top-$p$ at $T{=}1.5$ reach $p<10^{-70}$. Scores come from LongProc's own rules, so no judge is involved.}
\label{tab:longproc}
\small
\begin{tabular}{lrrrrrr}
\toprule
& \multicolumn{2}{c}{$T{=}0.7$} & \multicolumn{2}{c}{$T{=}1$} & \multicolumn{2}{c}{$T{=}1.5$} \\
\cmidrule(lr){2-3} \cmidrule(lr){4-5} \cmidrule(lr){6-7}
Sampler & Score & W/L & Score & W/L & Score & W/L \\
\midrule
Top-$p$ 0.9 & 31.1 &  & 29.1 &  & 13.4 &  \\
Top-$k$ 50 & 28.7 & 202--277$^{*}$ & 24.3 & 186--331$^{*}$ & 7.6 & 99--233$^{*}$ \\
Top-H 0.4 & 31.0 & 211--200 & 31.0 & 268--205$^{*}$ & 14.7 & 209--169$^{*}$ \\
Min-$p$ 0.1 & 30.3 & 222--253 & 28.5 & 246--240 & 23.9 & 399--100$^{*}$ \\
P-less & 31.0 & 215--203 & 31.2 & 274--193$^{*}$ & 30.8 & 532--73$^{*}$ \\
Top-$n\sigma$ 1 & 29.8 & 212--257$^{*}$ & 29.3 & 251--252 & 28.5 & 469--74$^{*}$ \\
Pure sampling & 28.6 & 202--278$^{*}$ & 22.7 & 169--361$^{*}$ & 2.7 & 31--302$^{*}$ \\
\bottomrule
\end{tabular}
\end{table}

\begin{figure}[h]
\centering
\includegraphics[width=\textwidth]{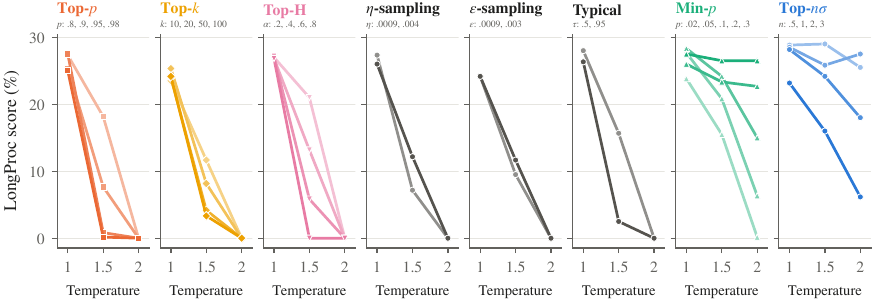}
\caption{Twenty-seven settings from eight rule families on LongProc, including $\eta$-sampling, $\epsilon$-sampling and locally typical sampling (Qwen3.5-4B and 9B, five tasks, 60 items per task). Lighter to darker lines follow the listed values. This repeats Figure~\ref{fig:map} with deterministic scoring and three more rule families. By $T{=}2$ every setting of every tail-admitting rule has collapsed, while top-$n\sigma$ with $n{\le}1$ and min-$p$ with $p{\ge}0.2$ have not.}
\label{fig:lphp}
\end{figure}

\begin{figure}[h]
\centering
\includegraphics[width=\textwidth]{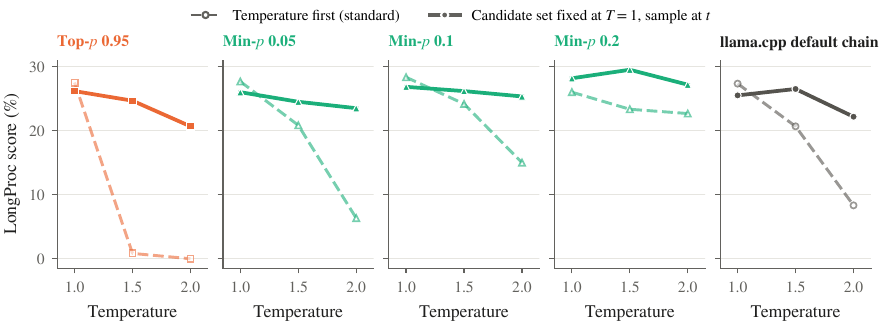}
\caption{The temperature-last test on LongProc with rule-based scoring. Dashed lines use the default order, and solid lines compute the candidate set at $T{=}1$ and then sample at $t$. The llama.cpp default chain is top-$k$ 40, top-$p$ 0.95, min-$p$ 0.05. Same models and items as Figure~\ref{fig:lphp}. As in Figure~\ref{fig:tl}, fixing the set keeps scores up where the default order collapses, this time with no judge at all, and it also rescues the llama.cpp default chain.}
\label{fig:lptl}
\end{figure}

Figure~\ref{fig:lphp} repeats the hyperparameter sweep of Section~\ref{sec:map} on the benchmark with deterministic scoring, and adds three  rule families. Every setting of top-$p$, top-$k$, Top-H, $\eta$-sampling, $\epsilon$-sampling and typical sampling scores 0\% or close to it at $T{=}2$. Top-$n\sigma$ 0.5 and 1 and min-$p$ 0.3 hold 25--28\%. Figure~\ref{fig:lptl} repeats the causal test of Section~\ref{sec:ablation}. Picking the candidate set at $T{=}1$ rescues top-$p$ 0.95 (21\% at $t{=}2$ vs.\ 0\%), min-$p$ at every $p$, and the llama.cpp default chain (22\% vs.\ 8\%).

\begin{table}[h]
\centering
\caption{LongBench v2, 503 questions, six models (Llama-3.1-8B, Nemotron-3-Nano-30B-A3B, Qwen3.5-4B, 9B, 35B-A3B, Qwen3.6-27B), contexts middle-truncated to 120K tokens, official two-stage chain-of-thought protocol (sampled reasoning up to 1,024 tokens, greedy answer extraction). Accuracy (\%) and W/L against top-$p$ 0.9. Answers are multiple choice and scored by exact match. The tight rules lose nothing at $T{=}1.5$, and the top-$n\sigma$ lead even at $T{=}1$ has $p<10^{-3}$.}
\label{tab:lb2}
\small
\begin{tabular}{lrrrrrr}
\toprule
& \multicolumn{2}{c}{$T{=}0.7$} & \multicolumn{2}{c}{$T{=}1$} & \multicolumn{2}{c}{$T{=}1.5$} \\
\cmidrule(lr){2-3} \cmidrule(lr){4-5} \cmidrule(lr){6-7}
Sampler & Score & W/L & Score & W/L & Score & W/L \\
\midrule
Top-$p$ 0.9 & 38.8 &  & 36.6 &  & 25.6 &  \\
Top-$k$ 50 & 38.3 & 298--314 & 36.9 & 336--329 & 29.6 & 442--322$^{*}$ \\
Top-H 0.4 & 39.2 & 295--284 & 39.1 & 346--272$^{*}$ & 35.4 & 557--261$^{*}$ \\
Min-$p$ 0.1 & 38.7 & 287--292 & 39.2 & 365--288$^{*}$ & 36.0 & 579--265$^{*}$ \\
P-less & 39.9 & 312--280 & 39.9 & 366--268$^{*}$ & 39.8 & 667--239$^{*}$ \\
Top-$n\sigma$ 1 & 39.4 & 308--291 & 39.5 & 355--269$^{*}$ & 38.3 & 635--253$^{*}$ \\
Pure sampling & 38.4 & 300--314 & 37.5 & 335--308 & 22.3 & 236--337$^{*}$ \\
\bottomrule
\end{tabular}
\end{table}

\begin{figure}[h]
\centering
\includegraphics[width=\textwidth]{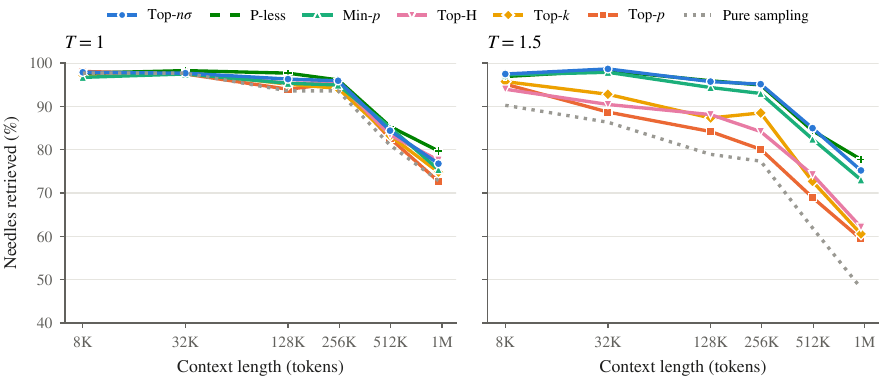}
\caption{Share of 16 key--value needles retrieved from PG-19 haystacks of 8K to one million tokens, four long-context models (Llama-3.1-8B-UltraLong-1M, Nemotron-3-Nano-30B-A3B, Qwen3.5-4B and 9B, the Qwen models with YaRN past 256K), 512 queries per point. Retrieval is scored by exact match, with no judge. At temperature 1 (left) the lines lie on top of each other, so any gap on the right comes from temperature and not from a difference in baseline ability. At 1.5 (right) the gap grows with context length. This is what we would expect if a longer context makes the model less sure of the answer and the tail-admitting rules turn that uncertainty into wrong digits. Table~\ref{tab:niah} has the paired tests.}
\label{fig:niah}
\end{figure}

\begin{table}[h]
\centering
\caption{Needle retrieval with all context lengths from 8K to 1M tokens pooled (five models, Qwen3.5-35B-A3B only up to 256K). Share of needles retrieved (\%) and W/L against top-$p$ 0.9 on identical queries. These are the paired tests behind Figure~\ref{fig:niah}.}
\label{tab:niah}
\small
\begin{tabular}{lrrrrrr}
\toprule
& \multicolumn{2}{c}{$T{=}0.7$} & \multicolumn{2}{c}{$T{=}1$} & \multicolumn{2}{c}{$T{=}1.5$} \\
\cmidrule(lr){2-3} \cmidrule(lr){4-5} \cmidrule(lr){6-7}
Sampler & Score & W/L & Score & W/L & Score & W/L \\
\midrule
Top-$p$ 0.9 & 92.9 &  & 91.4 &  & 82.3 &  \\
Top-$k$ 50 & 92.4 & 43--61 & 91.8 & 88--76 & 85.2 & 290--186$^{*}$ \\
Top-H 0.4 & 93.3 & 40--25 & 92.5 & 78--41$^{*}$ & 84.6 & 248--166$^{*}$ \\
Min-$p$ 0.1 & 92.9 & 38--39 & 92.1 & 75--50$^{*}$ & 91.1 & 385--70$^{*}$ \\
P-less & 93.6 & 47--23$^{*}$ & 93.5 & 90--15$^{*}$ & 92.6 & 402--35$^{*}$ \\
Top-$n\sigma$ 1 & 92.7 & 39--47 & 92.7 & 80--35$^{*}$ & 92.4 & 408--46$^{*}$ \\
Pure sampling & 92.1 & 35--62$^{*}$ & 90.9 & 72--91 & 77.4 & 143--321$^{*}$ \\
\bottomrule
\end{tabular}
\end{table}

\section{Forced-length replication in numbers}
\label{app:r1}

\begin{table}[h]
\centering
\caption{Forced 16K-token generation. Ten models, eight fiction prompts, seed 1, 80 generations per cell. Loop is the share of generations ending in a terminal loop. First and Last are the judge's overall score (1--10) on the first and final 1.5K-token windows. Read Loop together with First. At $T{=}1$ almost every rule loops, at $T{=}2$ the tail-admitting rules stop looping only because their text is word salad from the start (First near 1), and top-$n\sigma$ keeps its $T{=}1$ scores.}
\label{tab:r1}
\small
\begin{tabular}{lccccccccc}
\toprule
 & \multicolumn{3}{c}{$T{=}1.0$} & \multicolumn{3}{c}{$T{=}1.5$} & \multicolumn{3}{c}{$T{=}2.0$} \\
\cmidrule(lr){2-4} \cmidrule(lr){5-7} \cmidrule(lr){8-10}
Sampler & Loop & First & Last & Loop & First & Last & Loop & First & Last \\
\midrule
Pure sampling & 0.41 & 4.0 & 2.5 & 0.03 & 1.1 & 1.0 & 0.00 & 1.0 & 1.0 \\
Top-$k$ 50 & 0.56 & 5.2 & 2.7 & 0.00 & 3.2 & 1.5 & 0.00 & 2.0 & 1.2 \\
Top-$p$ 0.9 & 0.95 & 5.2 & 2.5 & 0.05 & 1.5 & 1.2 & 0.00 & 1.0 & 1.0 \\
Min-$p$ 0.1 & 0.97 & 5.2 & 2.5 & 0.64 & 5.3 & 2.4 & 0.09 & 3.8 & 1.6 \\
Top-H 0.4 & 0.97 & 4.8 & 2.6 & 0.20 & 4.5 & 2.2 & 0.00 & 1.8 & 1.1 \\
\pless{} & 0.99 & 4.8 & 2.6 & 0.99 & 5.2 & 2.6 & 0.44 & 3.6 & 1.9 \\
Top-$n\sigma$ 1 & 1.00 & 5.2 & 2.5 & 0.90 & 5.2 & 2.5 & 0.71 & 5.1 & 2.4 \\
\midrule
Greedy & \multicolumn{3}{c}{1.00 / 4.8 / 2.5} &  &  &  &  &  &  \\
\bottomrule
\end{tabular}

\end{table}

\begin{table}[h]
\centering
\caption{LiveCodeBench v6, Qwen3.8-27B in thinking mode, 175 problems $\times$ 2 samples, 32K tokens, official tests in a sandbox. W/L counts problem-samples the rule solved and top-$p$ 0.95 didn't at the same $T$, and the reverse. $^{*}$ marks a sign test with $p<0.05$. Every rule ties up to $T{=}1$, and at $T{=}1.3$ only top-$p$ loses ground.}
\label{tab:lcb}
\small
\begin{tabular}{lcccccc}
\toprule
 & \multicolumn{2}{c}{$T{=}0.6$} & \multicolumn{2}{c}{$T{=}1$} & \multicolumn{2}{c}{$T{=}1.3$} \\
\cmidrule(lr){2-3} \cmidrule(lr){4-5} \cmidrule(lr){6-7}
Sampler & pass@1 & W/L & pass@1 & W/L & pass@1 & W/L \\
\midrule
Top-$p$ 0.95 & 62.0 & base & 60.9 & base & 50.6 & base \\
Top-$n\sigma$ 1 & 63.4 & 14/9 & 63.7 & 22/12 & 63.4 & 53/8$^{*}$ \\
Min-$p$ 0.1 & 62.6 & 14/12 & 62.9 & 21/14 & 63.1 & 50/6$^{*}$ \\
\pless{} & 62.0 & 14/14 & 62.6 & 22/16 & 62.6 & 49/7$^{*}$ \\
Top-H 0.4 & 61.7 & 15/16 & 62.6 & 22/16 & 62.9 & 50/7$^{*}$ \\
\bottomrule
\end{tabular}

\end{table}

\begin{table}[h]
\centering
\caption{LiveCodeBench v6, Qwen3.6-27B in thinking mode, same protocol as Table~\ref{tab:lcb}. This model runs out of its 32K-token budget on 44--53\% of samples in every cell, which makes it a hard case for every rule. At $T{=}0.6$, \pless{} and Top-H lose to top-$p$, and at $T{=}1.3$ top-$n\sigma$ still beats top-$p$ ($p{=}0.02$).}
\label{tab:lcb36}
\small
\begin{tabular}{lcccccc}
\toprule
 & \multicolumn{2}{c}{$T{=}0.6$} & \multicolumn{2}{c}{$T{=}1$} & \multicolumn{2}{c}{$T{=}1.3$} \\
\cmidrule(lr){2-3} \cmidrule(lr){4-5} \cmidrule(lr){6-7}
Sampler & pass@1 & W/L & pass@1 & W/L & pass@1 & W/L \\
\midrule
Top-$p$ 0.95 & 52.0 & base & 49.1 & base & 46.3 & base \\
Top-$n\sigma$ 1 & 52.9 & 20/17 & 51.4 & 18/10 & 50.3 & 24/10$^{*}$ \\
Min-$p$ 0.1 & 49.4 & 16/25 & 51.1 & 15/8 & 48.6 & 21/13 \\
\pless{} & 46.3 & 14/34$^{*}$ & 45.4 & 12/25$^{*}$ & 49.7 & 24/12 \\
Top-H 0.4 & 46.9 & 16/34$^{*}$ & 48.6 & 22/24 & 49.7 & 26/14 \\
\bottomrule
\end{tabular}

\end{table}

\section{Candidate sets and the temperature-last ablation in numbers}
\label{app:numbers}

\paragraph{Figure~\ref{fig:teaser}(a).} Llama-3.1-8B-Instruct decodes the science-fiction prompt greedily for 512 tokens. Out of the steps whose top probability is in $[0.3,0.7]$, we show the one with the median entropy (step 329, context ``\ldots It had been launched over three hundred years ago,'', $p_{\max}{=}0.45$, $H{=}1.43$ nats). Candidate-set sizes are computed exactly over the full 128K-token vocabulary.

\begin{figure}[h]
\centering
\includegraphics[width=\textwidth]{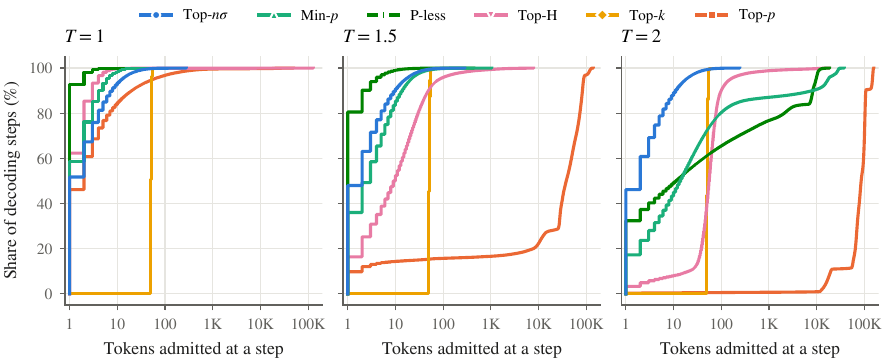}
\caption{Cumulative share of decoding steps that admit at most $x$ tokens, over the first 2,048 tokens of the ten-model forced-length generations. The further left a curve rises, the smaller that rule's candidate sets. At $T{=}2$ top-$n\sigma$ admits 10 tokens or fewer on 89\% of steps, close to its 94\% at $T{=}1$. Min-$p$ and P-less stay small on half of all steps but pick up a long tail of steps with thousands of candidates, which medians hide and which fits their moderate losses at $T{=}2$. Top-H goes past 50 tokens on most steps, and top-$p$ admits tens of thousands on nearly every step.}
\label{fig:pooldist}
\end{figure}

\begin{table}[h]
\centering
\caption{Median candidate-set size per decoding step over the first 2,048 tokens of the Table~\ref{tab:r1} generations (ten models). Singleton is the share of steps at $T{=}1$ where only one token is admitted. Compare the columns within each row. Only top-$n\sigma$ (by construction) and top-$k$ (fixed $k$) don't grow with temperature.}
\label{tab:pool}
\small
\begin{tabular}{lcccc}
\toprule
Sampler & $T{=}1.0$ & $T{=}1.5$ & $T{=}2.0$ & Singleton at $T{=}1$ \\
\midrule
Pure sampling & all & all & all & 0.00 \\
Top-$k$ 50 & 51 & 51 & 51 & 0.00 \\
Top-$p$ 0.9 & 2 & 41,508 & 89,581 & 0.46 \\
Min-$p$ 0.1 & 1 & 3 & 14 & 0.59 \\
Top-H 0.4 & 1 & 10 & 54 & 0.62 \\
\pless{} & 1 & 1 & 21 & 0.93 \\
Top-$n\sigma$ 1 & 1 & 1 & 1 & 0.52 \\
\bottomrule
\end{tabular}

\end{table}

\begin{table}[h]
\centering
\caption{Mean judged window quality (1--10), four models, 32 forced 8K-token generations per cell. ``Temperature first'' is the default order (Figure~\ref{fig:map}). ``Candidate set fixed at $T{=}1$'' truncates on the untempered distribution and then samples at temperature $t$ inside the kept set (Appendix~\ref{app:defs}). Where the fixed-set columns match the $T{=}1$ column, temperature stopped hurting once the set could no longer grow. That happens for Top-H, min-$p$ and \pless{} and not for top-$p$ or top-$k$.}
\label{tab:tl}
\small
\begin{tabular}{lccccccc}
\toprule
 & $T{=}1$ & \multicolumn{3}{c}{Temperature first} & \multicolumn{3}{c}{Candidate set fixed at $T{=}1$} \\
\cmidrule(lr){3-5} \cmidrule(lr){6-8}
Sampler & & $T{=}1.6$ & $T{=}2$ & $T{=}3$ & $t{=}1.5$ & $t{=}2$ & $t{=}3$ \\
\midrule
Top-$p$ 0.9 & 4.7 & 1.0 & 1.0 & 1.0 & 3.4 & 2.7 & 2.6 \\
Top-$k$ 50 & 4.8 & 1.9 & 1.4 & 1.3 & 2.0 & 1.4 & 1.2 \\
Top-H 0.4 & 4.2 & 2.5 & 1.3 & 1.2 & 4.2 & 4.5 & 4.5 \\
Min-$p$ 0.1 & 4.6 & 4.4 & 2.6 & 1.1 & 4.5 & 4.6 & 4.6 \\
\pless{} & 4.2 & 4.6 & 2.9 & 1.1 & 4.2 & 4.2 & 4.1 \\
\bottomrule
\end{tabular}

\end{table}

\begin{table}[h]
\centering
\caption{Published long-writing benchmarks, i.e. Arena-Write and LongBench-Write (the numbers behind Figure~\ref{fig:bench}) and WritingBench critic scores (1--10, six models). The ordering is the same on all three benchmarks.}
\label{tab:bench}
\small
\resizebox{\textwidth}{!}{\begin{tabular}{lcccccccccccc}
\toprule
 & \multicolumn{3}{c}{Arena-Write healthy} & \multicolumn{3}{c}{Arena-Write last window} & \multicolumn{3}{c}{LongBench-Write $S$} & \multicolumn{3}{c}{WritingBench} \\
\cmidrule(lr){2-4} \cmidrule(lr){5-7} \cmidrule(lr){8-10} \cmidrule(lr){11-13}
Sampler & $T{=}1$ & $T{=}1.5$ & $T{=}2$ & $T{=}1$ & $T{=}1.5$ & $T{=}2$ & $T{=}1$ & $T{=}1.5$ & $T{=}2$ & $T{=}1$ & $T{=}1.5$ & $T{=}2$ \\
\midrule
Pure sampling & 0.78 & 0.14 & 0.00 & 6.4 & 2.0 & 1.0 & 78.6 & 32.1 & 15.0 & 6.85 & 3.71 & 1.26 \\
Top-$k$ 50 & 0.93 & 0.39 & 0.15 & 6.8 & 4.5 & 2.7 & 79.8 & 67.4 & 43.1 & 6.94 & 6.48 & 5.00 \\
Top-$p$ 0.9 & 0.92 & 0.23 & 0.01 & 6.8 & 2.6 & 1.2 & 80.8 & 42.4 & 21.1 & 6.89 & 4.37 & 2.20 \\
Min-$p$ 0.1 & 0.92 & 0.95 & 0.69 & 6.8 & 6.9 & 5.8 & 80.3 & 81.3 & 74.9 & 6.88 & 6.97 & 6.67 \\
Top-H 0.4 & 0.90 & 0.72 & 0.10 & 6.6 & 5.9 & 2.3 & 78.8 & 76.6 & 37.8 & 6.84 & 6.81 & 4.35 \\
\pless{} & 0.90 & 0.90 & 0.74 & 6.7 & 6.7 & 6.1 & 78.4 & 79.6 & 77.1 & 6.84 & 6.86 & 6.78 \\
Top-$n\sigma$ 1 & 0.93 & 0.93 & 0.93 & 6.8 & 6.8 & 6.8 & 79.7 & 80.4 & 80.6 & 6.88 & 6.90 & 6.92 \\
\bottomrule
\end{tabular}
}
\end{table}

\section{Robustness, sampling stacks and context depth in numbers}
\label{app:robust}

\subsection{Natural-length continuation and non-fiction}
\label{sec:natural}

Forcing a model to keep going after it wants to stop is an artificial stress, so we also ran a protocol much closer to how long outputs get produced in practice. The model can emit end-of-sequence whenever it wants, and every time it stops we add a ``Continue.'' turn, up to a 16K-token budget with 4K-token turns (Table~\ref{tab:r4}, nine of the ten models, gpt-oss-120b excluded).

\begin{figure}[h]
\centering
\includegraphics[width=\textwidth]{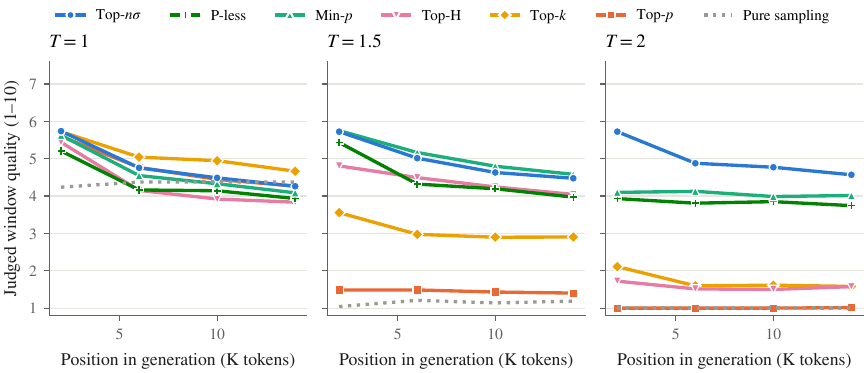}
\caption{Natural-length continuation, position by position. The model can stop whenever it wants and is told to continue (``Continue.'') up to 16K tokens. Nine models, judged 1.5K-token windows grouped into 4K-token bins. This is closer to real use than forced length. At $T{=}2$ top-$n\sigma$ opens at 5.7 and stays above 4.5 to the end while top-$p$ scores 1.0 in every bin, the same split as in Figure~\ref{fig:hero}, so that split is not an artifact of forcing the length.}
\label{fig:natural}
\end{figure}

 At $T{=}1$ the final judged window goes from about 2.5 under forced length to 3.8--4.5. Top-$p$'s loop rate drops from 0.95 to 0.50 and top-$n\sigma$'s from 1.00 to 0.39. Quality at $T{=}1$ is still close across rules (final-window means 3.8--4.5). \pless{} loops the most at $T{\le}1.5$ (0.89 and 0.76), which fits with its near-greedy candidate set. Turning up the temperature brings back the forced-length split. At $T{=}2$ top-$p$, top-$k$, Top-H and pure sampling score 1.0--2.1 on their first window, min-$p$ and \pless{} drop to 4.0--4.1, and top-$n\sigma$ holds 5.7, though it still loops in 26\% of generations.

Non-fiction looks the same (Appendix~\ref{app:nonfic}, six models). At $T{=}1$ every rule ends at 3.3--3.6 on the final window. At $T{=}1.5$ top-$p$ and top-$k$ drop to first-window scores of 3.1 and 4.4, while top-$n\sigma$ and min-$p$ hold theirs (6.3 to 6.6 and 6.3 to 6.5).

\subsection{Second seed, newer models, second judge}
\label{app:robust2}

\begin{figure}[h]
\centering
\includegraphics[width=\textwidth]{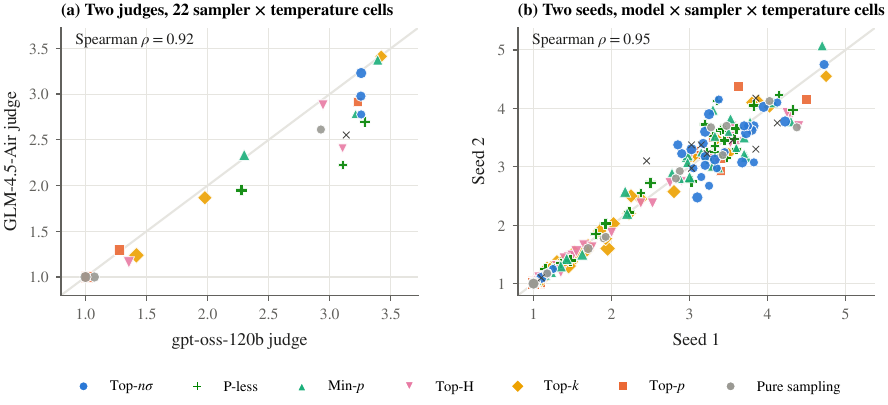}
\caption{(a) Cell means from gpt-oss-120b against GLM-4.5-Air on the same 8,796 windows. (b) Per-model cell means for seed 1 against seed 2 (all ten models, 8 generations per cell per seed). Bigger markers mean higher temperature. Points on the diagonal mean the two judges, or the two seeds, agree. Both clouds hug it, so neither the choice of judge nor the random seed drives the forced-length conclusions.}
\label{fig:agree}
\end{figure}

\begin{figure}[h]
\centering
\includegraphics[width=\textwidth]{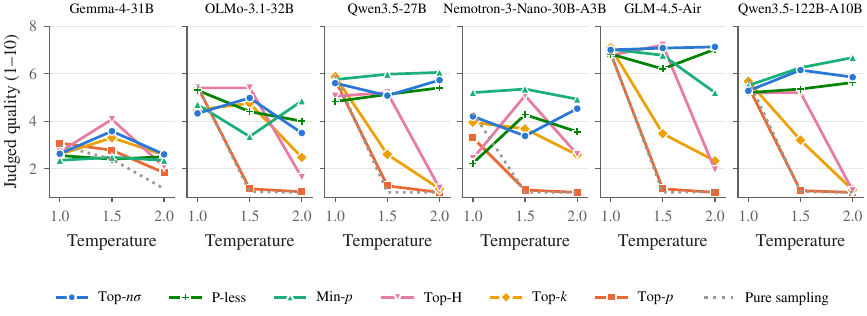}
\caption{Six newer models, two of them mixture-of-experts models up to 122B parameters. Mean judged window quality under forced 16K-token fiction, 8 generations per cell. On every model except Gemma-4-31B, whose top-$p$ set stays near one token (Appendix~\ref{app:gemma}), top-$p$ and pure sampling collapse by $T{=}1.5$ while min-$p$, P-less and top-$n\sigma$ stay well above them. The effect does not go away at larger scale or with mixture-of-experts models.}
\label{fig:large}
\end{figure}

We ran three more checks on the forced-length results. A second seed on all ten models (80 generations per cell, Table~\ref{tab:r9}, Appendix~\ref{app:robust}) reproduces Table~\ref{tab:r1}. At $T{=}1$ top-$p$, min-$p$, Top-H, \pless{} and top-$n\sigma$ loop in 96--100\% of generations, and at $T{=}2$ top-$n\sigma$ opens at 5.3 and top-$p$ at 1.0. Six newer models (Gemma-4-31B, OLMo-3.1-32B, Qwen3.5-27B, Nemotron-3-Nano-30B-A3B, and the mixture-of-experts GLM-4.5-Air and Qwen3.5-122B-A10B, Table~\ref{tab:r10}) loop less at $T{=}1$ (56--81\% for the five truncation rules), but they split the same way at $T{=}2$, where top-$n\sigma$ opens at 7.4 and top-$p$ at 1.7. Last, we had a second judge from a different family, GLM-4.5-Air, rescore all 8,796 windows of the ten-model grid. The two judges agree at Spearman $\rho{=}0.74$ on single windows and $\rho{=}0.92$ on the 22 sampler-by-temperature cell means, so the cell-level conclusions don't depend on which judge we use.

\begin{table}[h]
\centering
\caption{Second seed of the forced-length replication, all ten models, 80 generations per cell. Columns as in Table~\ref{tab:r1}, which it closely reproduces.}
\label{tab:r9}
\small
\begin{tabular}{lccccccccc}
\toprule
 & \multicolumn{3}{c}{$T{=}1.0$} & \multicolumn{3}{c}{$T{=}1.5$} & \multicolumn{3}{c}{$T{=}2.0$} \\
\cmidrule(lr){2-4} \cmidrule(lr){5-7} \cmidrule(lr){8-10}
Sampler & Loop & First & Last & Loop & First & Last & Loop & First & Last \\
\midrule
Pure sampling & 0.46 & 3.9 & 2.4 & 0.00 & 1.1 & 1.1 & 0.00 & 1.0 & 1.0 \\
Top-$k$ 50 & 0.55 & 5.3 & 2.5 & 0.01 & 3.2 & 1.6 & 0.00 & 1.9 & 1.1 \\
Top-$p$ 0.9 & 0.96 & 5.2 & 2.5 & 0.04 & 1.5 & 1.1 & 0.00 & 1.0 & 1.0 \\
Min-$p$ 0.1 & 0.97 & 5.1 & 2.5 & 0.66 & 5.3 & 2.6 & 0.10 & 3.9 & 1.5 \\
Top-H 0.4 & 1.00 & 5.0 & 2.4 & 0.21 & 4.4 & 2.0 & 0.00 & 1.8 & 1.1 \\
\pless{} & 0.99 & 4.8 & 2.8 & 0.99 & 5.1 & 2.7 & 0.39 & 3.6 & 1.9 \\
Top-$n\sigma$ 1 & 0.99 & 5.2 & 2.3 & 0.89 & 5.3 & 2.4 & 0.76 & 5.3 & 2.2 \\
\midrule
Greedy & \multicolumn{3}{c}{1.00 / 4.7 / 2.7} &  &  &  &  &  &  \\
\bottomrule
\end{tabular}

\end{table}

\begin{table}[h]
\centering
\caption{Forced-length grid on six newer models (Gemma-4-31B, OLMo-3.1-32B, Qwen3.5-27B, Nemotron-3-Nano-30B-A3B, GLM-4.5-Air, Qwen3.5-122B-A10B), 48 generations per cell. These models loop less at $T{=}1$ than the core ten but split the same way at $T{=}2$.}
\label{tab:r10}
\small
\begin{tabular}{lccccccccc}
\toprule
 & \multicolumn{3}{c}{$T{=}1.0$} & \multicolumn{3}{c}{$T{=}1.5$} & \multicolumn{3}{c}{$T{=}2.0$} \\
\cmidrule(lr){2-4} \cmidrule(lr){5-7} \cmidrule(lr){8-10}
Sampler & Loop & First & Last & Loop & First & Last & Loop & First & Last \\
\midrule
Pure sampling & 0.62 & 7.1 & 4.1 & 0.00 & 2.1 & 1.1 & 0.00 & 1.1 & 1.0 \\
Top-$k$ 50 & 0.50 & 7.0 & 4.1 & 0.10 & 6.7 & 2.7 & 0.00 & 3.4 & 1.5 \\
Top-$p$ 0.9 & 0.69 & 7.0 & 4.2 & 0.17 & 2.6 & 1.1 & 0.00 & 1.7 & 1.0 \\
Min-$p$ 0.1 & 0.56 & 7.3 & 4.3 & 0.54 & 7.4 & 4.2 & 0.21 & 7.2 & 4.0 \\
Top-H 0.4 & 0.81 & 6.7 & 3.7 & 0.40 & 7.3 & 4.2 & 0.00 & 2.8 & 1.5 \\
\pless{} & 0.75 & 6.8 & 3.5 & 0.88 & 7.3 & 3.5 & 0.65 & 7.1 & 3.7 \\
Top-$n\sigma$ 1 & 0.73 & 7.1 & 3.9 & 0.67 & 7.0 & 4.0 & 0.69 & 7.4 & 3.8 \\
\midrule
Greedy & \multicolumn{3}{c}{0.81 / 7.1 / 3.5} &  &  &  &  &  &  \\
\bottomrule
\end{tabular}

\end{table}

\begin{table}[h]
\centering
\caption{Sampling stacks on eight models, Llama-3.3-70B and gpt-oss-120b included (forced 16K fiction). Loop is the terminal-loop rate and Quality is mean judged window quality. The ``none'' rows are the bare rules on the same eight models. DRY cuts loops at $T{=}1$ without raising quality, and the top-2 floor only matters at high temperature (Appendix~\ref{sec:stack}).}
\label{tab:stack}
\small
\begin{tabular}{llcccccc}
\toprule
 & & \multicolumn{2}{c}{$T{=}1$} & \multicolumn{2}{c}{$T{=}1.5$} & \multicolumn{2}{c}{$T{=}2$} \\
\cmidrule(lr){3-4} \cmidrule(lr){5-6} \cmidrule(lr){7-8}
Rule & Stack & Loop & Quality & Loop & Quality & Loop & Quality \\
\midrule
Top-$p$ 0.9 & none & 0.95 & 3.2 & 0.06 & 1.3 & 0.00 & 1.0 \\
 & + DRY & 0.38 & 3.4 & 0.00 & 1.2 & 0.00 & 1.0 \\
 & + top-2 floor & 0.88 & 3.3 & 0.05 & 1.4 & 0.00 & 1.0 \\
\addlinespace[2pt]
Min-$p$ 0.1 & none & 0.97 & 3.1 & 0.64 & 3.3 & 0.11 & 2.4 \\
 & + DRY & 0.34 & 3.2 & 0.22 & 3.5 & 0.05 & 2.4 \\
 & + top-2 floor & 0.97 & 3.0 & 0.11 & 3.4 & 0.00 & 2.4 \\
\addlinespace[2pt]
Top-H 0.4 & none & 0.97 & 3.1 & 0.23 & 2.9 & 0.00 & 1.4 \\
 & + DRY & 0.36 & 3.2 & 0.08 & 3.0 & 0.00 & 1.4 \\
 & + top-2 floor & 0.97 & 3.0 & 0.14 & 3.0 & 0.00 & 1.3 \\
\addlinespace[2pt]
\pless{} & none & 1.00 & 3.0 & 0.98 & 3.3 & 0.45 & 2.4 \\
 & + DRY & 0.31 & 3.1 & 0.36 & 3.1 & 0.17 & 2.5 \\
 & + top-2 floor & 0.97 & 3.2 & 0.20 & 3.2 & 0.05 & 2.5 \\
\addlinespace[2pt]
Top-$n\sigma$ 1 & none & 1.00 & 3.3 & 0.91 & 3.2 & 0.75 & 3.2 \\
 & + DRY & 0.27 & 3.5 & 0.31 & 3.2 & 0.39 & 3.4 \\
 & + top-2 floor & 0.92 & 3.1 & 0.16 & 3.3 & 0.02 & 3.3 \\
\bottomrule
\end{tabular}

\end{table}

\begin{table}[h]
\centering
\caption{Continuing PG-19 novels from increasing context depth, 8K generated tokens. Quality gap is the mean judged quality of tail-excluding rules minus that of tail-admitting rules. Entropy is the mean per-step entropy (nats) of the untempered distribution over the first 2K generated tokens at $T{=}1$. The gap is near zero at $T{\le}1$ at every depth and shows no trend with depth, so reading a longer context does not by itself make the sampler matter (Appendix~\ref{sec:depth}).}
\label{tab:depth}
\small
\begin{tabular}{lcccc}
\toprule
Context depth & 2K & 16K & 64K & 128K \\
\midrule
Quality gap, $T{=}1$ & -0.03 & +0.08 & -0.37 & -0.08 \\
Quality gap, $T{=}1.3$ & +1.21 & +1.43 & +1.21 & +1.41 \\
Quality gap, $T{=}1.6$ & +1.50 & +1.51 & +1.58 & +1.48 \\
\midrule
Entropy at $T{=}1$, \texttt{gemma4-12b} & 0.12 & 0.10 & 0.19 & -- \\
Entropy at $T{=}1$, \texttt{llama3.1-8b} & 2.91 & 2.42 & 1.66 & -- \\
Entropy at $T{=}1$, \texttt{nemotron3-30b-a3b} & 1.18 & 1.11 & 1.08 & 1.32 \\
Entropy at $T{=}1$, \texttt{qwen3.5-27b} & 0.94 & 1.09 & 1.32 & -- \\
Entropy at $T{=}1$, \texttt{qwen3.5-35b-a3b} & 1.16 & 1.32 & 1.44 & 1.70 \\
Entropy at $T{=}1$, \texttt{qwen3.5-9b} & 1.49 & 1.38 & 1.65 & 1.86 \\
\bottomrule
\end{tabular}

\end{table}

\begin{table}[h]
\centering
\caption{Natural-length multi-turn continuation (``Continue.'' after each stop, 16K-token budget), seven models, eight fiction genres, about 42 generations per cell. Columns as in Table~\ref{tab:r1}. Letting the model stop and continue halves the loop rate of several rules, and it still doesn't separate them at $T{=}1$.}
\label{tab:r4}
\small
\begin{tabular}{lccccccccc}
\toprule
 & \multicolumn{3}{c}{$T{=}1.0$} & \multicolumn{3}{c}{$T{=}1.5$} & \multicolumn{3}{c}{$T{=}2.0$} \\
\cmidrule(lr){2-4} \cmidrule(lr){5-7} \cmidrule(lr){8-10}
Sampler & Loop & First & Last & Loop & First & Last & Loop & First & Last \\
\midrule
Pure sampling & 0.04 & 4.2 & 4.4 & 0.03 & 1.0 & 1.2 & 0.00 & 1.0 & 1.0 \\
Top-$k$ 50 & 0.21 & 5.7 & 4.5 & 0.00 & 3.6 & 2.7 & 0.00 & 2.1 & 1.6 \\
Top-$p$ 0.9 & 0.50 & 5.6 & 4.2 & 0.01 & 1.5 & 1.4 & 0.00 & 1.0 & 1.0 \\
Min-$p$ 0.1 & 0.64 & 5.6 & 4.0 & 0.14 & 5.8 & 4.5 & 0.00 & 4.1 & 3.9 \\
Top-H 0.4 & 0.75 & 5.5 & 3.8 & 0.11 & 4.8 & 4.0 & 0.01 & 1.7 & 1.6 \\
\pless{} & 0.89 & 5.2 & 3.9 & 0.76 & 5.5 & 3.9 & 0.25 & 4.0 & 3.7 \\
Top-$n\sigma$ 1 & 0.39 & 5.7 & 4.2 & 0.32 & 5.7 & 4.4 & 0.26 & 5.7 & 4.5 \\
\midrule
Greedy & \multicolumn{3}{c}{0.88 / 5.2 / 3.8} &  &  &  &  &  &  \\
\bottomrule
\end{tabular}

\end{table}

\section{Stress-test prompts}
\label{app:prompts}

\paragraph{Fiction template.} \emph{Write a \{genre\} novel based on the following premise. Premise: \{premise\} Write the complete novel, chapter by chapter, with full scenes, dialogue, and character development. It should be very long (at least 30,000 words). Do not summarize, do not skip ahead, and do not stop early.}

\paragraph{Fiction premises (one per genre).}
\begin{itemize}[leftmargin=*,nosep]
\item \textbf{science fiction.} A generation ship's maintenance engineer discovers that the ship's navigation AI has been quietly altering course for three hundred years.
\item \textbf{fantasy.} A lighthouse keeper on the edge of the known world must keep a magical flame burning while the sea itself begins to forget its shape.
\item \textbf{mystery.} A retired crossword compiler is asked to solve the murder of a rival whose final puzzle names the killer.
\item \textbf{horror.} A family moves into a farmhouse where every clock runs seven minutes fast, and the seven minutes are filled with someone else's footsteps.
\item \textbf{literary fiction.} Three estranged siblings return to their late mother's seaside house to divide her belongings and discover her unsent letters.
\item \textbf{romance.} Two competing food-truck owners are forced to share a parking spot for an entire summer festival season.
\item \textbf{thriller.} A bank's night-shift security analyst detects a robbery in progress that the bank's executives insist is not happening.
\item \textbf{historical fiction.} In 1850s San Francisco, a Chinese herbalist and an Irish dockworker form an unlikely partnership during the Gold Rush.
\end{itemize}

\paragraph{Non-fiction prompts.} Each is followed by: \emph{It should be very long (at least 30,000 words). Do not summarize, do not skip ahead, and do not stop early.}
\begin{itemize}[leftmargin=*,nosep]
\item Write a comprehensive technical book titled 'Designing Distributed Databases from First Principles'. Cover consensus, replication, partitioning, transactions, storage engines, and failure handling, with worked examples in every chapter.
\item Write a detailed narrative history of the Byzantine Empire from the founding of Constantinople to its fall in 1453, organized into chapters by era.
\item Write a popular-science book explaining how the human immune system works, from first principles to modern immunotherapy, with a chapter for each major topic.
\item Write a book-length philosophical treatise on the nature of personal identity over time, engaging with thought experiments and objections in every chapter.
\item Write a complete field guide and natural history of temperate rainforests, covering geology, climate, plants, fungi, animals, and conservation, chapter by chapter.
\item Write a complete hands-on programming textbook that teaches compilers by building one in Python, from lexing to code generation, with code and explanations in every chapter.
\item Write a book-length biography of a fictional 19th-century inventor named Adelaide Marsh, covering her childhood, inventions, rivalries, and legacy.
\item Write a long travel memoir about walking the length of the Danube river from its source to the Black Sea, describing every stage of the journey.
\end{itemize}

\section{Non-fiction at 16K}
\label{app:nonfic}

\begin{figure}[h]
\centering
\includegraphics[width=\textwidth]{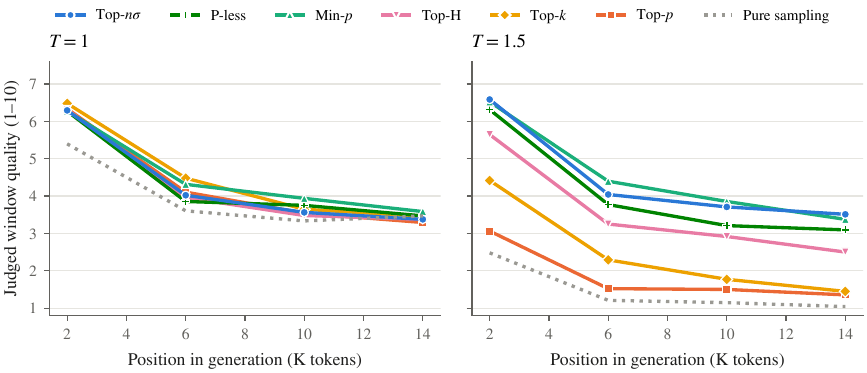}
\caption{Non-fiction under forced 16K-token generation, position by position (six models, eight prompts, 4K-token bins). The split at $T{=}1.5$ matches fiction (Figure~\ref{fig:hero}). $T{=}2$ was not run for non-fiction.}
\label{fig:nonfic}
\end{figure}

This is the forced-length protocol of Table~\ref{tab:r1} run on eight non-fiction prompts and six models (Llama-3.2-1B, Qwen2.5-3B, Llama-3.1-8B, Qwen3.5-9B, Gemma-4-31B, Llama-3.3-70B).

\begin{table}[h]
\centering
\caption{Forced 16K-token non-fiction, six models, eight prompts. Columns as in Table~\ref{tab:r1}. At $T{=}1.5$ top-$p$ and top-$k$ open far lower than at $T{=}1$, while top-$n\sigma$ and min-$p$ hold their scores.}
\label{tab:nonfic}
\small
\begin{tabular}{lcccccc}
\toprule
 & \multicolumn{3}{c}{$T{=}1.0$} & \multicolumn{3}{c}{$T{=}1.5$} \\
\cmidrule(lr){2-4} \cmidrule(lr){5-7}
Sampler & Loop & First & Last & Loop & First & Last \\
\midrule
Pure sampling & 0.56 & 5.4 & 3.4 & 0.00 & 2.5 & 1.1 \\
Top-$k$ 50 & 0.69 & 6.5 & 3.3 & 0.06 & 4.4 & 1.4 \\
Top-$p$ 0.9 & 0.83 & 6.3 & 3.3 & 0.21 & 3.1 & 1.4 \\
Min-$p$ 0.1 & 0.88 & 6.3 & 3.6 & 0.69 & 6.5 & 3.4 \\
Top-H 0.4 & 0.90 & 6.3 & 3.4 & 0.31 & 5.6 & 2.6 \\
\pless{} & 0.88 & 6.3 & 3.5 & 0.85 & 6.3 & 3.0 \\
Top-$n\sigma$ 1 & 0.85 & 6.3 & 3.4 & 0.75 & 6.6 & 3.5 \\
\midrule
Greedy & \multicolumn{3}{c}{0.92 / 6.2 / 3.4} &  &  &  \\
\bottomrule
\end{tabular}

\end{table}

\section{Limitations}
\label{sec:limitations}

\paragraph{Judge.}
Every quality score comes from one LLM judge (gpt-oss-120b) grading 1.5K-token windows, and LLM judges are known to be biased by length, style and position~\citep{zheng2023judging}. That is why we also report loop rates, the realized candidate set and objective SWE-bench test outcomes, so that no conclusion depends on the judge alone. We did not run a human evaluation.

\paragraph{Coverage.}
The replication covers ten open-weight models from 1B to 120B parameters. The agentic study uses one agent harness, with most grids on a 9B model and replications on three larger Qwen3.5 models and two Gemma-4 models. We can't test closed models because their providers don't let users control the sampler. Each forced-length cell has 80 generations, which isn't enough to resolve differences of a few tenths of a judge point between rules that don't collapse, and we don't claim any.

\paragraph{Forced length.}
Forcing generation to a fixed length is an artificial stress. Natural-length continuation (Section~\ref{sec:natural}) lowers loop rates and leaves the conclusions the same.

\end{document}